\documentclass[journal]{IEEEtran}
\usepackage{cite}

\usepackage{booktabs}
\usepackage{multirow}
\ifCLASSINFOpdf
\usepackage[pdftex]{graphicx}
\else
\fi
\usepackage{amsmath}
\usepackage{bm}
\usepackage{amsfonts,amssymb,textcomp}

\ifCLASSOPTIONcompsoc
  \usepackage[caption=false,font=normalsize,labelfont=sf,textfont=sf]{subfig}
\else
  \usepackage[caption=false,font=footnotesize]{subfig}
\fi

\usepackage[linkcolor=green, citecolor=green, anchorcolor=green]{hyperref}

\begin{document}
%
\title{CFC-Net: A Critical Feature Capturing Network for Arbitrary-Oriented Object Detection in Remote Sensing Images}
%
%
%

\author{Qi~Ming,
        Lingjuan~Miao,
        Zhiqiang~Zhou,
        and~Yunpeng~Dong

}
\maketitle

\begin{abstract}
Object detection in optical remote sensing images is an important and challenging task.  In recent years, the methods based on convolutional neural networks have made good progress. However, due to the large variation in object scale, aspect ratio, and arbitrary orientation, the detection performance is difficult to be further improved. In this paper, we discuss the role of discriminative features in object detection, and then propose a Critical Feature Capturing Network (CFC-Net) to improve detection accuracy from three aspects: building powerful feature representation, refining preset anchors, and optimizing label assignment. Specifically, we first decouple the classification and regression features, and then construct robust critical features adapted to the respective tasks through the Polarization Attention Module (PAM). With the extracted discriminative regression features, the Rotation Anchor Refinement Module (R-ARM) performs localization refinement on preset horizontal anchors to obtain superior rotation anchors. Next, the Dynamic Anchor Learning (DAL) strategy is given to adaptively select high-quality anchors based on their ability to capture critical features. The proposed framework creates more powerful semantic representations for objects in  remote sensing images and achieves high-performance real-time object detection. Experimental results on three remote sensing datasets including HRSC2016, DOTA, and UCAS-AOD show that our method achieves superior detection performance compared with many state-of-the-art approaches. Code and models are available at https://github.com/ming71/CFC-Net.
\end{abstract}


%
\IEEEpeerreviewmaketitle

\section{Introduction}
%
%
%
%
\IEEEPARstart{O}{bject}  detection in optical remote sensing images is a vital computer vision technique which aims at classifying and locating objects in remote sensing images. It is widely used in crop monitoring, resource exploration, environmental monitoring, military reconnaissance, etc. With the explosive growth of available remote sensing data, identifying objects of interest from massive amounts of remote sensing images has gradually become a challenging task. Most of the traditional methods use handcrafted features to identify objects \cite{li2012automatic, han2014efficient, han2014object, zhu2010novel, eikvil2009classification}. Although much progress has been made, there are still problems such as low-efficiency, insufficient robustness, and poor performance.

\begin{figure}[t]
	\centering
	\subfloat[]{	
		\includegraphics[width=0.2\textwidth]{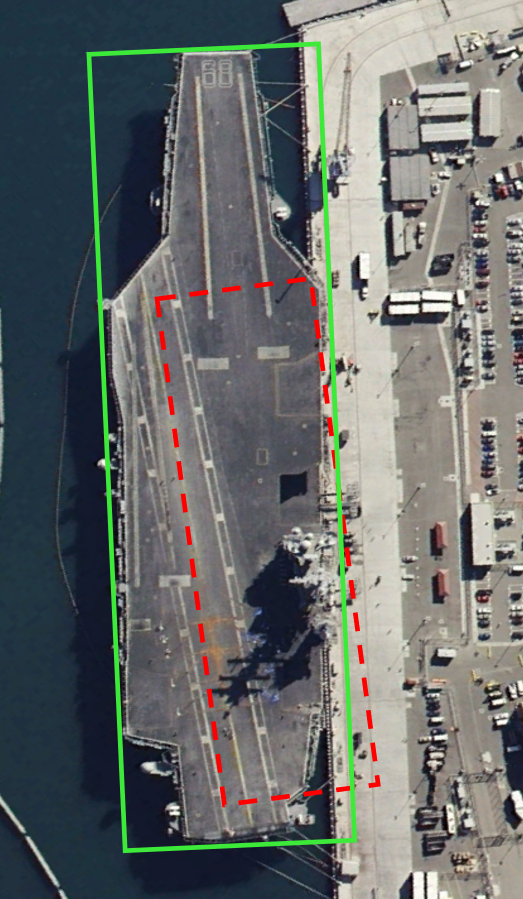}}\hspace{1mm}
	\subfloat[]{	
		\includegraphics[width=0.2\textwidth]{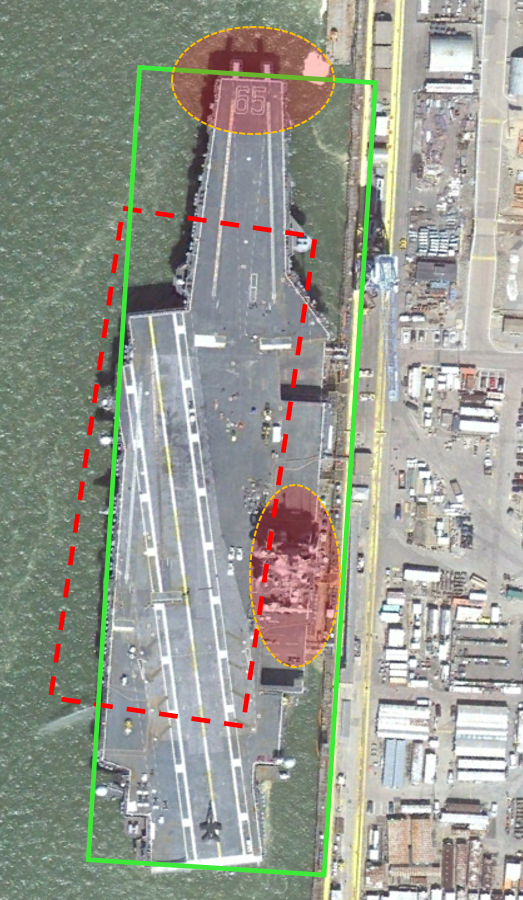}}
	\caption{Illustration of the role of critical features in classification task. Predicted bounding boxes  (green) are regressed from predefined anchor boxes (red). The ground truth classes of (a) and (b) are marked as A and B respectively, while the predicted object categories are all A. Only the anchors that capture the critical features required to identify the object (such as island and bow here) can achieve the correct classification prediction.}
	\label{Fig1}
\end{figure} 

In recent years, the development of convolution neural networks (CNNs) has greatly improved the performance of object detection. Most CNN-based detection frameworks first extract features through convolution operation, and then preset a series of prior boxes (anchors) on the feature maps. Subsequently, classification and regression will be performed on these anchors to obtain the bounding boxes of objects. The powerful ability to automatically extract features of CNN makes it possible to achieve efficient object detection on massive images. Currently, the CNN-based models have been widely used in the object detection in remote sensing images, such as road detection \cite{yang2019road}, vehicle detection \cite{ji2019vehicle}, airport detection \cite{liu2019multi}, and ship detection \cite{wu2018inshore, li2018hsf}. 

Although CNN-based approaches have made good progress, they are often directly derived from generic object detection frameworks. It is difficult for these methods to detect objects with a wide variety of scales, aspect ratios, and orientations in remote sensing images. For example, the orientation of objects varies greatly  in remote sensing imagery, while the mainstream generic detectors utilize predefined horizontal anchors to predict these rotated ground-truth (GT) boxes. The horizontal boxes often contain a lot of background which may mislead the detection. There are some approaches that use rotated anchors to locate arbitrary-oriented objects \cite{liu2018arbitrary, zhang2018toward, liu2017rotated, ding2019learning, li2020novel, liao2018rotation, fu2020rotation}. But it is hard for rotation anchors to achieve good spatial alignment with GT boxes, and they can not ensure to provide sufficiently good semantic information for classification and regression. 

Some recent researches address the above problems by designing more powerful feature representations \cite{cheng2016learning, zhou2017oriented, deng2018multi, wang2019fmssd, fu2020rotation, liao2018rotation}. However, they only focus on a certain type of characteristics of remote sensing targets, such as rotation invariant features\cite{cheng2016learning, zhou2017oriented}and scale sensitive features\cite{deng2018multi, wang2019fmssd}. They cannot automatically extract and utilize more complex and discriminative features. Another commonly used method is to manually set a large number of anchors covering different aspect ratios, scales, and orientations to achieve better spatial alignment with targets. In this way, sufficient high-quality anchors can be obtained and better performance can be achieved. Nevertheless, excessive preset anchors bring about three problems: (1) Most anchors are backgrounds that cannot be used for bounding box regression, which leads to severely redundant calculation. (2) The parameters of the prior anchors need to be careful manually set, otherwise, they would not obtain good alignment with GT boxes. (3) There are a large number of low-quality negative samples in the excessive laid anchors which are not conducive to network convergence. The above-mentioned issues lead to the fact that densely preset anchors are still unable to effectively handle the difficulties of remote sensing object detection.

\begin{figure}[t]
	\centering
	\subfloat[]{
		\includegraphics[width=0.2\textwidth]{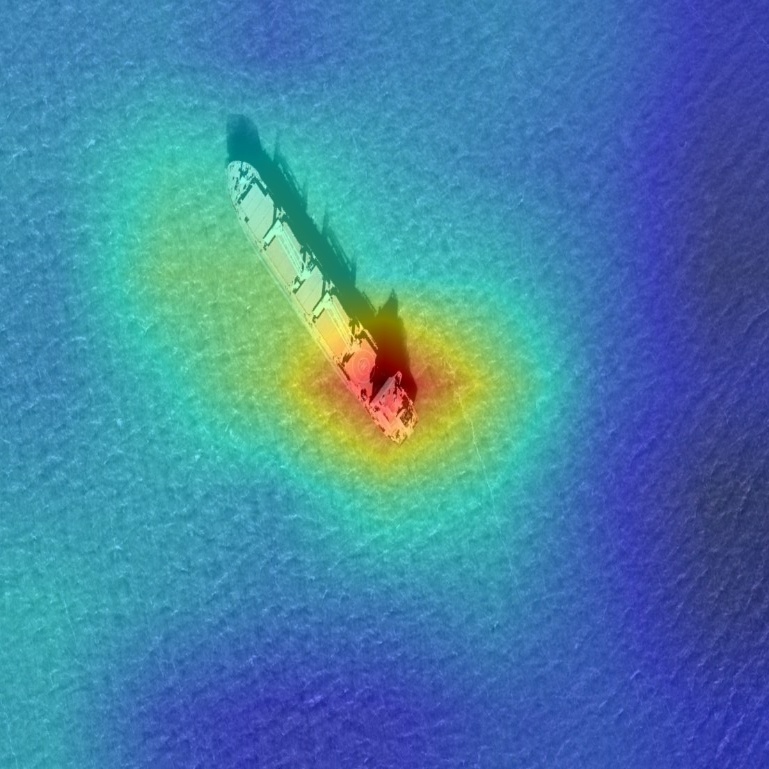}}\hspace{0mm}
	\quad
	\subfloat[]{
		\includegraphics[width=0.2\textwidth]{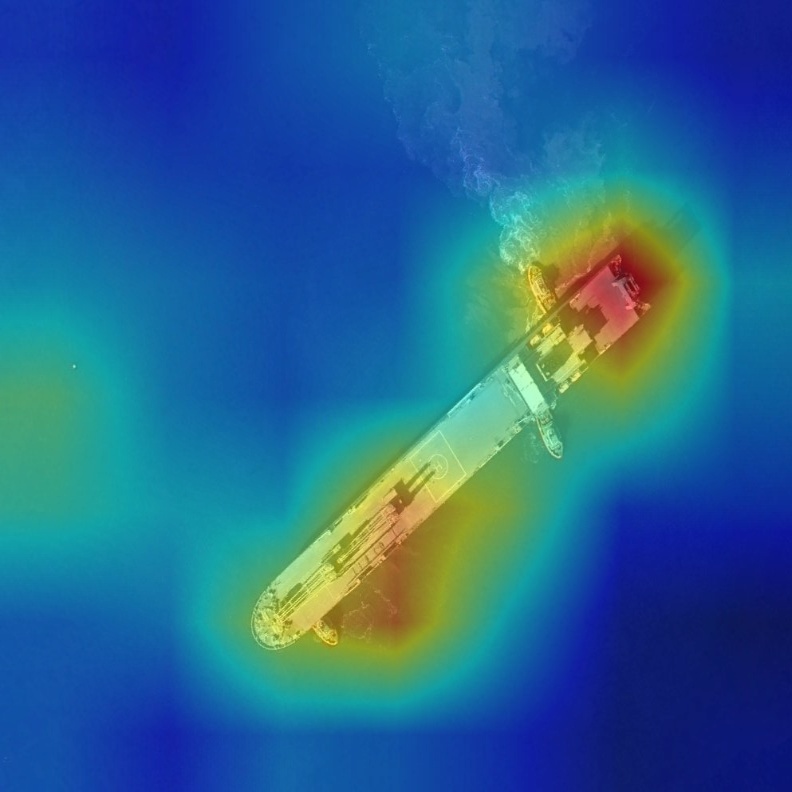}}

	\subfloat[]{
		\includegraphics[width=0.2\textwidth]{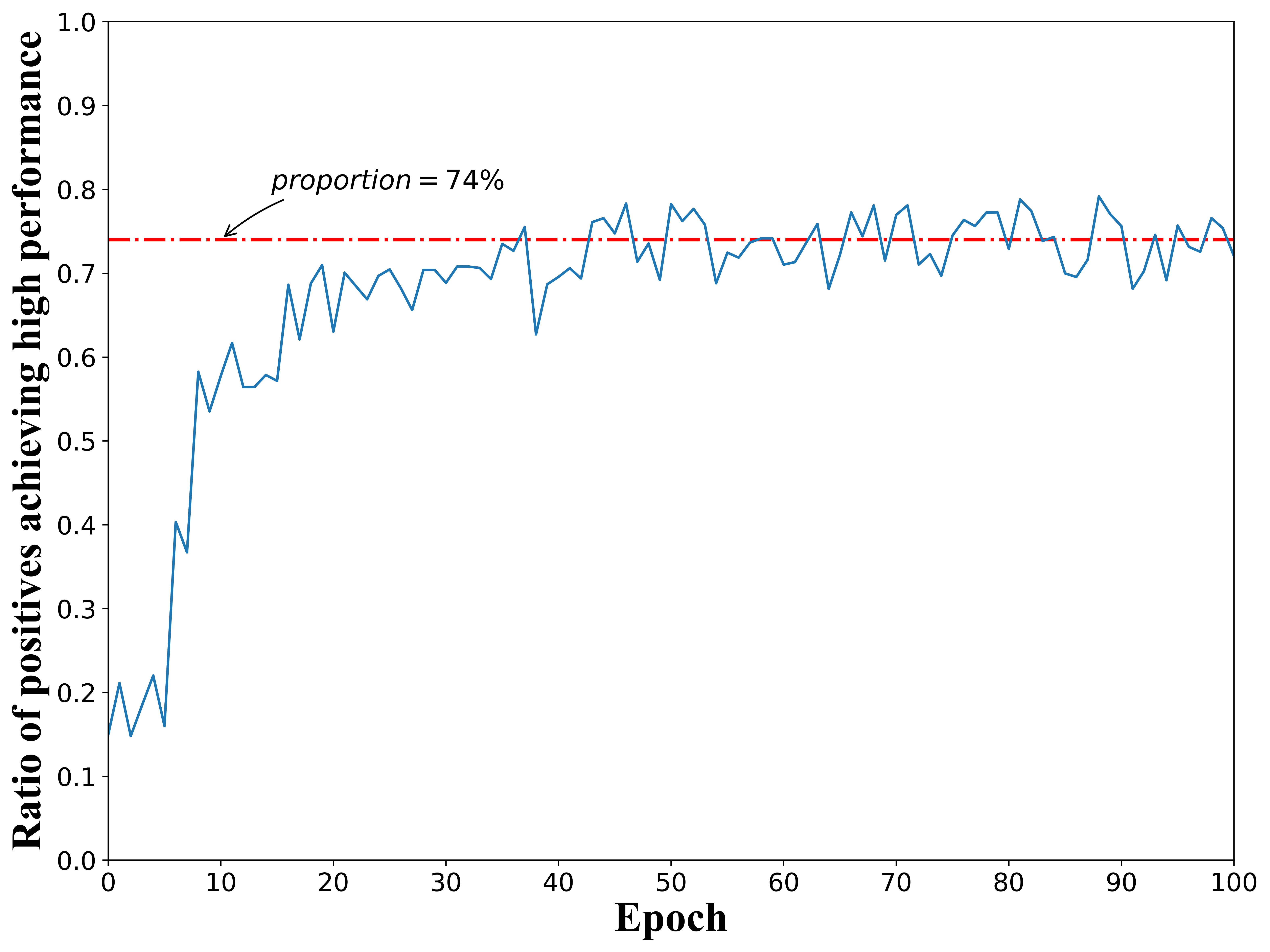}}\hspace{0mm}
	\quad
	\subfloat[]{
		\includegraphics[width=0.2\textwidth]{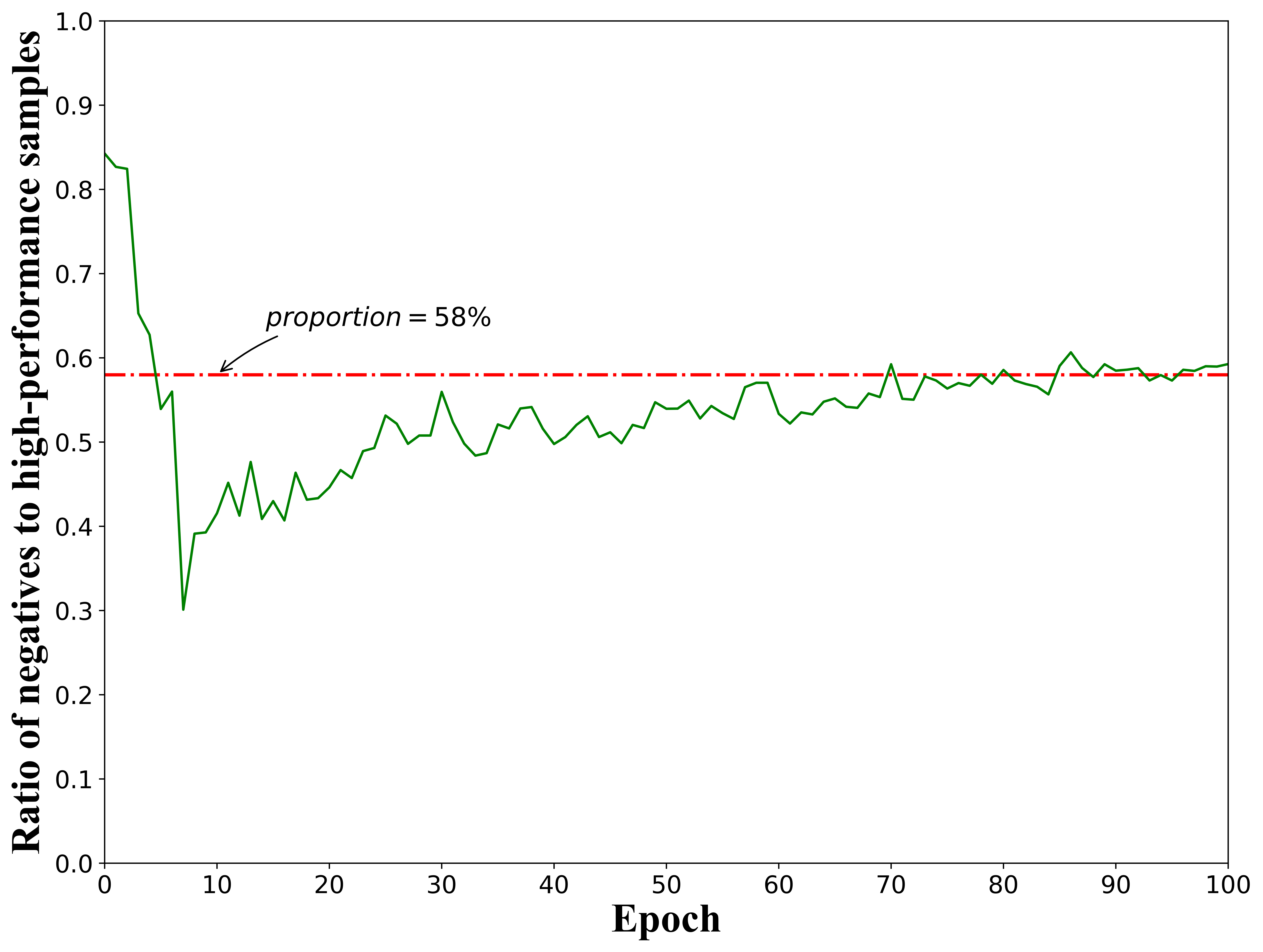}}
	\caption{Analysis of the importance of critical features. (a)-(b) Discriminative feature activation map in object detection. (c) The proportion of positive samples with high-quality detections among all positives. (d) The proportion of high-quality detections that regressed from negatives.}
	\label{Fig2}
\end{figure}

To figure out how the complex variabilities of remote sensing objects make it difficult to achieve high-performance detection, in this paper we introduce the essential concept named critical features, which indicates discriminative features required for accurate classification or localization. Taking the classification task as an example, most anchor-based detectors treat the anchors in Fig.~\ref{Fig1}(a) and Fig.~\ref{Fig1}(b) as positive samples, that is, the IoU between these anchors and GT boxes is higher than 0.5. But the anchor in Fig.~\ref{Fig1}(b) does not capture the discriminative features of the island and bow which are necessary to identify the ship B. Although this anchor achieves accurate localization, it leads to incorrect classification results, thereby degrading detection performance. Furthermore, by visualizing the features extracted by CNN, it is found that the critical features required to identify objects for classification and regression are not always evenly distributed on the object, but may be on local areas such as the bow and stern (see Fig.~\ref{Fig2}(a) and Fig.~\ref{Fig2}(b)). The preset anchors need to capture these critical features to achieve accurate detection. This is similar to the conclusion of some previous work \cite{li2016novel,wu2018inshore}. However, the mainstream rotation detectors are more likely to select anchors with high IoU with GT boxes as positives, but ignore high-quality anchors that contain critical features, which eventually leads to the unstable training process and poor performance. The distribution of the localization ability for all candidates is shown in Fig.~\ref{Fig2}(c) to support this viewpoint. It can be seen that only 74\% of positive anchors can achieve high-quality detection (with output IoU larger than 0.5) after regression, which indicates that even the positive anchors still cannot guarantee precise localization. We attribute this phenomenon to the fact that some of the selected positives do not capture the critical features required by the regression task. Besides, as shown in Fig.~\ref{Fig2}(d), surprisingly more than half of the anchors (about 58\% in this case) that achieve accurate detection are regressed from samples that are divided as negatives. It means that a large number of negative anchors capture the critical features well but have not been effectively utilized at all. The inconsistency between the training sample division and the regression results will further lead to a gap between the classification scores and localization accuracy of the detections. Based on the above observations, we conclude that one of the key issues in object detection in remote sensing imagery is whether the anchors can capture the critical features of the objects.

In this paper, based on the viewpoint of the significance of critical features discussed above, the Critical Feature Capturing Network (CFC-Net) is proposed to achieve high-performance object detection in optical remote sensing imagery. Specifically, CFC-Net first uses a well-designed Polarization Attention Module (PAM) to generate different feature pyramids for classification and regression tasks, and then we can obtain task-specific critical features that are more discriminative as well as easy to be captured. Next, the Rotation Anchor Refinement Module (R-ARM) refines the preset horizontal anchors to better capture the regression critical features to obtain high-quality rotation anchors. Finally, in the training process, the Dynamic Anchor Learning (DAL) strategy is adopted to select the high-quality anchors that capture critical features as positives to ensure superior detection performance after training. Due to the proper construction and utilization of critical features, CFC-Net achieves the state-of-the-art detection performance using only one anchor, which makes it became a both high-performance and memory-saving method. The code is available to facilitate future research.

The contributions of this article are summarized as follows:
\begin{enumerate}
\item We point out the existence of critical features through experiments, and interpret common challenges for object detection in remote sensing imagery from this perspective.
\item A novel object detection framework CFC-Net is proposed to extract the critical features and utilize high-quality anchors that capture the critical features to achieve superior detection performance. 
\item Polarized attention is proposed to construct task-specific critical features. Decoupled critical features provide more useful semantic information for individual tasks, which is beneficial to accurate classification and regression.
\item The dynamic anchor selection strategy selects high-quality anchors that capture the critical regression features to bridge the inconsistency between classification and regression, and thus greatly improves the performance of detection.
\end{enumerate}

The rest of this article is organized as follows. Section \uppercase\expandafter{\romannumeral2} introduces the related work of object detection. Section \uppercase\expandafter{\romannumeral3} elaborates on the proposed method. Section \uppercase\expandafter{\romannumeral4} shows the experimental results and analysis. Finally, conclusions are drawn in Section \uppercase\expandafter{\romannumeral5}.

\section{Related work}
Object detection in remote sensing images  has a wide range of application scenarios and has been receiving extensive attention in recent years. Most of the early traditional methods  use handcraft features to detect remote sensing objects, such as shape and texture features \cite{li2012automatic, zhu2010novel, eikvil2009classification}, scale-invariant features \cite{han2014efficient}, and saliency \cite{han2014object}. For instance, Zhu \textit{et al.} \cite{zhu2010novel} achieves accurate ship detection based on shape and texture features. Eikvil \textit{et al.} \cite{eikvil2009classification} utilizes spatial geometric properties and gray level features for vehicle detection in satellite images. These approaches have achieved satisfactory performance for specific scenes, but their low efficiency and poor generalization make it hard to detect objects in complex scenarios. 

Recently, with the great success of convolution neural networks, generic object detection has been strongly promoted. Mainstream CNN-based object detection methods can be classified into two categories: one-stage detectors and two-stage detectors. The two-stage detectors first generate a series of proposals, and then perform classification and regression on these regions to obtain the detection results \cite{girshick2014rich, girshick2015fast, ren2015faster}. These algorithms usually have high accuracy but slow inference speed. The one-stage detectors, such as the YOLO series \cite{redmon2016you, redmon2017yolo9000, redmon2018yolov3} and SSD \cite{liu2016ssd}, directly conduct classification and regression on the prior anchors without region proposal generation. Compared with the two-stage detectors, one-stage methods have relatively low accuracy, but are faster and can achieve real-time object detection.

Deep learning methods have been widely used in object detection in remote sensing images. A series of CNN-based approaches have been proposed and achieved good performance. However, some methods are directly developed from the generic object detection framework \cite{deng2018multi, zhang2016weakly}, which detect objects with horizontal bounding box. It is hard for the horizontal box to distinguish densely arranged remote sensing targets and is prone to misdetection. To solve this problem, some studies introduced an additional orientation dimension to achieve the oriented object detection \cite{liu2018arbitrary, zhang2018toward, liu2017rotated}. For example,  Liu \textit{et al.} \cite{liu2018arbitrary} integrates the angle regression into the YOLOv2 \cite{redmon2017yolo9000} to detect rotated ships. R$^2$PN \cite{zhang2018toward} detects rotated ships by generating oblique region of interest (RoI). RR-CNN \cite{liu2017rotated} uses the rotated RoI pooling layer, which makes the RoI feature better aligned with the orientation of the object to ensure accurate detection. However, in order to have a higher overlap with the rotated objects, these methods preset densely arranged rotation anchors. Most of the anchors have no intersection with the targets, which brings a lot of redundant computation and the severe imbalance problem. Some work alleviates the issue by setting fewer anchors but still maintaining detection performanc\cite{ding2019learning, yang2019r3det}. RoI Transformer \cite{ding2019learning} adopts horizontal anchors to learn the rotated RoI through spatial transformation, and thus a few horizontal anchors work well for oriented object detection. R$^3$Det \cite{ yang2019r3det} achieves state-of-the-art performance through cascade regression and feature alignment is performed on horizontal anchors. Despite the success of these methods, it is still difficult for horizontal anchors to match the rotation objects and the number of preset anchors is still large. Different from the previous work, our CFC-Net uses only one anchor for faster inference and achieves high-quality rotation object detection.

There are also some methods trying to construct better feature representation to alleviate the difficulty of anchor matching caused by large scale, shape, and orientation variations \cite{zhou2017oriented, cheng2016learning, wang2019fmssd, li2020novel, zhang2019cad, fu2020rotation}. For instance,  ORN \cite{zhou2017oriented} performs feature extraction through the rotated convolution kernel to achieve rotation invariance. RICNN \cite{cheng2016learning} optimizes the feature representation by learning a  rotation-invariant layer. FMSSD \cite{wang2019fmssd} aggregates the context information in different scales to cope with the multi-scale objects in large-scale remote sensing imagery. Li \textit{et al.}  \cite{li2020novel}  proposed a shape-adaptive pooling to  extract the features of the ships with various aspect ratios, and then multilevel features are incorporated to generate compact feature representation for ship detection. RRD \cite{liao2018rotation} observes that shared features degrade performance due to the incompatibility of the classification and regression tasks, and thus the rotation-invariant and rotation-sensitive features are constructed for classification and regression tasks, respectively. But these work only pays attention to a certain aspect of the object characteristics, and cannot comprehensively cover the discriminative features required for object detection. According to the proposed concept of critical features, we believe that the detection performance depends on whether the prior anchors effectively capture these critical features, not limited to the rotation-invariant features or scale-invariant features. Therefore, the clear and easy-to-capture powerful critical feature representation is very important for object detection. The proposed CFC-Net extracts and utilizes task-sensitive critical features for classification and regression tasks respectively so that the detector obtains substantial performance improvements from the more discriminative critical feature representation.

\begin{figure*}[t]
	\centering
	\includegraphics[width=1.0\textwidth]{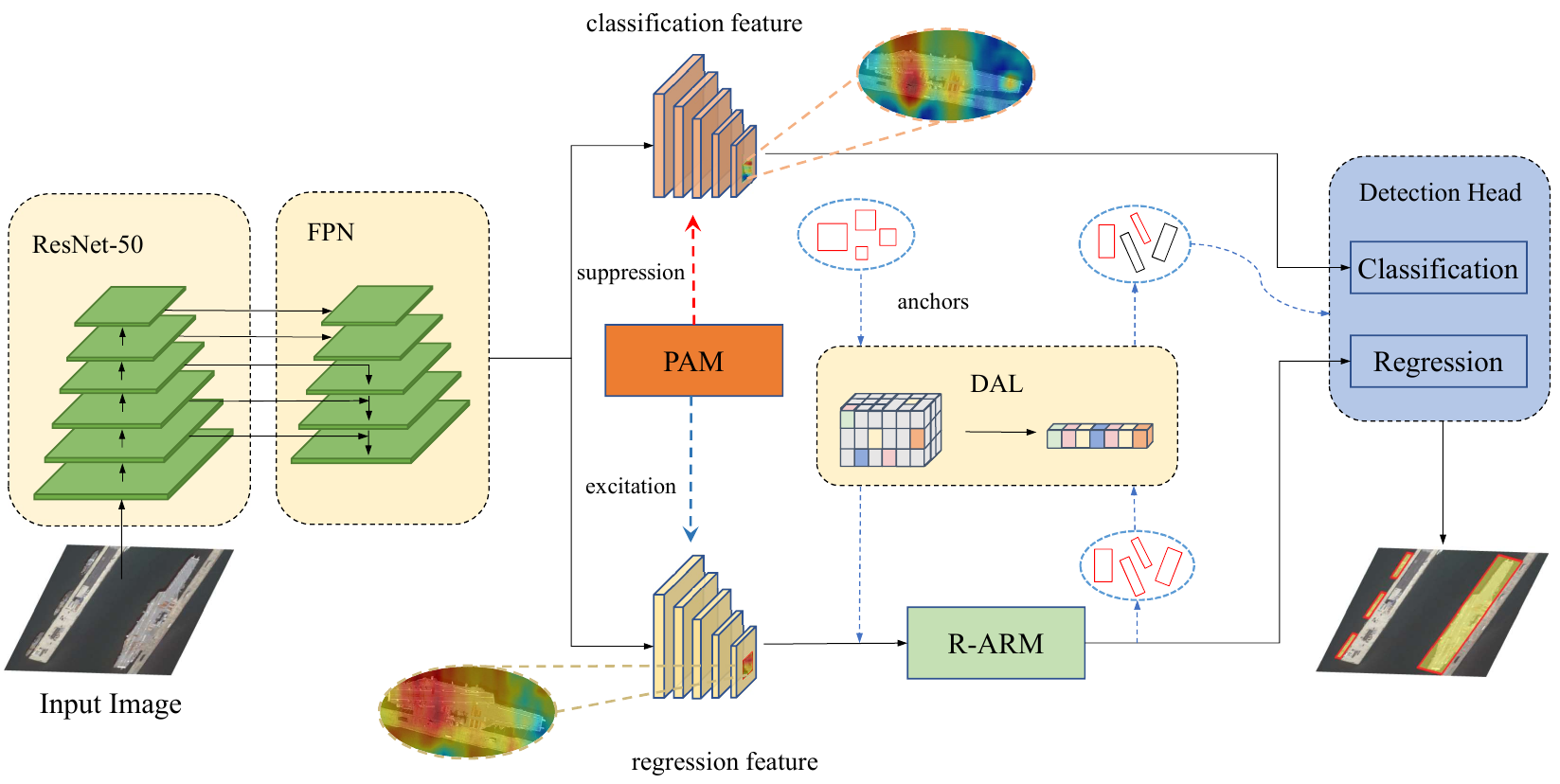} 
	\caption{Framework of the proposed CFC-Net.}
	\label{Fig3}
\end{figure*}

\section{Proposed Method}
The overall structure of CFC-Net is shown in Fig.~\ref{Fig3}. It uses ResNet-50 as the backbone network. Firstly, we build multi-scale feature pyramids through FPN \cite{lin2017feature}, and then the decoupled features that are sensitive to classification and regression are generated through the proposed PAM. Subsequently, anchor refinement is conducted via R-ARM to obtain the high-quality rotation candidates based on the critical regression features. Finally, through the DAL strategy, anchors that capture critical features are dynamically selected as positive samples for training. In this way, the inconsistency between classification and regression can be alleviated and thus the detection performance can be effectively improved. The details of the proposed CFC-Net are elaborated below.

\subsection{Polarization Attention Module}

\begin{figure*}[t]
	\centering
	\includegraphics[width=1.0\textwidth]{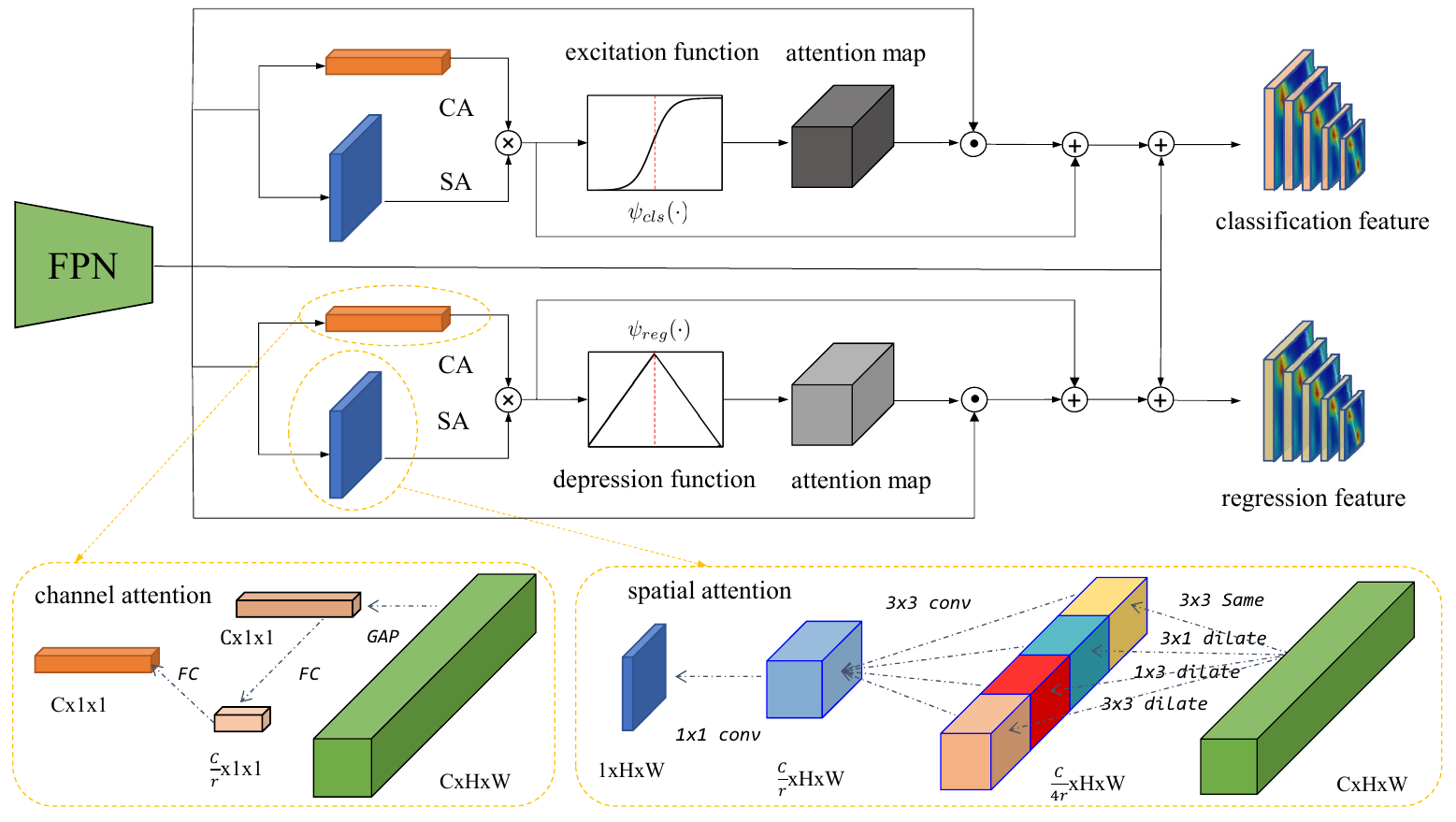} 
	\caption{Illustration of the PAM module. SA denotes spatial attention, CA represents channel-wise attention.}
	\label{Fig4}
\end{figure*}

In most object detection frameworks, both classification and regression rely on the shared features. However, as mentioned in \cite{liao2018rotation, song2020revisiting}, the shared features degrade performance owing to the incompatibility between the two tasks. For example, the regression branch of detectors needs to be sensitive to change of the angle so as to achieve accurate orientation prediction, while classification branch is supposed to have the same response to different angles. Therefore, rotation-invariant features are beneficial to classification task, but it is not conducive to bounding box regression.

We propose Polarization Attention Module (PAM) to avoid the feature interference between different tasks and effectively extract the task-specific critical features. The overall structure of PAM is shown in Fig.~\ref{Fig4}. Firstly, we build separate feature pyramids for different tasks, which is called dual FPN. Next, a well-designed polarization attention mechanism is applied to enhance the representation ability of features. Through the polarization function, different branches generate the discriminative features required for respective tasks. Specifically, for classification, we tend to select high-response global features to reduce noise interference. For regression, we pay more attention to the features of object boundaries and suppress the influence of irrelevant high activation regions. 

Given input feature $\mathbf{F} \in \mathbb{R}^{C \times H \times W}$, we construct task-sensitive features as follows:
\begin{equation}
\begin{aligned}
\mathbf{M}&=\mathbf{M}_{\mathbf{c}}(\mathbf{F}) \otimes \mathbf{M}_{\mathbf{s}}(\mathbf{F}),  \\
\mathbf{F}^{\prime}&=\mathbf{M} + \psi(\sigma(\mathbf{M}) )\odot \mathbf{F} + \mathbf{F},
\label{Eq1}
\end{aligned}
\end{equation}
where $\otimes$ and $\odot$ represent tensor product and element-wise multiplication, respectively. $\sigma$ denotes sigmoid function. Firstly, we extract channel-wise attention map $\mathbf{M}_{\mathbf{c}}$ and spatial attention map $\mathbf{M}_{\mathbf{s}}$ from input features through convolution operations. The purpose of channel attention is to extract the channel-wise relationship of the feature maps. The weight of each channel is extracted by global average pooling and fully connected layers as:
\begin{equation}
\mathbf{M}_{\mathbf{c}}(\mathbf{F}) =\sigma\left(\mathbf{W}_{\mathbf{1}}\left(\mathbf{W}_{\mathbf{0}}\left(\mathbf{F}_{\mathbf{gap}}\right)\right)\right),
\label{Eq2}
\end{equation}
where $\mathbf{F}_{\mathbf{gap}}$ is obtained from input feature $\mathbf{F}$ via global average pooling, $\mathbf{W}_{\mathbf{0}} \in \mathbb{R}^{C / r \times C}$ and $\mathbf{W}_{\mathbf{1}} \in \mathbb{R}^{C \times C / r}$ are the weights of the fully connected layers. $\sigma$ represents the sigmoid function. 

Correspondingly, spatial attention is used to model the dependencies between pixels of the input image. It is computed as:
\begin{equation}
\mathbf{M}_{\mathbf{s}}(\mathbf{F}) =\sigma\left(c^{3\times 3}(cat((c^{3\times 3},c^{1\times 3}_d,c^{3\times 1}_d,c^{3\times 3}_d)(\mathbf{F})))\right),
\label{Eq3}
\end{equation}
in which $c^{3\times 3}$ represents convolution of a $3\times 3$ filter. $c^{1\times 3}_d,c^{3\times 1}_d,c^{3\times 3}_d$ respectively denote dilated convolution of different kernel sizes. $cat$ denotes concatenation of features. Dilated convolution is adopted here to expand the receptive field of the convolution kernels. At the same time, convolution kernels with different aspect ratios are used to better detect slender objects such as ships and bridges.

Next, the attention response map $\mathbf{M}$ for a specific task is obtained by multiplying the two attention maps. On this basis, we further build the powerful task-sensitive critical feature representation through the task-specific polarization function $\psi(\cdot)$. For classification, the features are expected to pay more attention to the high-response part on feature maps, and ignore the part of less important clues which may be used for localization or or may bring interference noise. We use the following excitation function to achieve the function:
\begin{equation}
\psi_{cls}(x) = \frac{1}{1+e^{-\eta(x-0.5)}},
\label{Eq4}
\end{equation}
where $\eta$ is the modulation factor used to control the intensity of feature activation (set to 15 in our experiment). Since the high-response area of critical classification features is enough to achieve accurate classification, there is no need to pursue too much information. Consequently, the effect of high-response critical classification features are excited, while irrelevant features with attention weight less than 0.5 are suppressed. In this way, the classifier is able to pay less attention to the difficult-to-classify areas and reduce the risk of overfitting and misjudgment.

Meanwhile, for the regression branch, the critical features are often scattered on the edges of object. We expect that the feature maps focus on as many visual clues as possible for object localization, such as object contours and contextual information. To this end, we use the following depression function to process the input features:
\begin{equation}
\psi_{reg}(x)=\left\{\begin{array}{ll}
x & \text { if } x<0.5, \\
1-x & \text { otherwise. }
\end{array}\right.
\label{Eq5}
\end{equation}
Different from the classification task, a strong response to a patch of the object edge is not conducive to locating the entire object. In Eq.(\ref{Eq5}), the depression function suppresses the area with the high response in the regression feature, which enforces the model to seek potential visual clues to achieve accurate localization. The curves of polarization function $\psi(\cdot)$ are shown in Fig.~\ref{Fig4}. 

\begin{figure*}[t]
	\centering
	\includegraphics[width=1.0\textwidth]{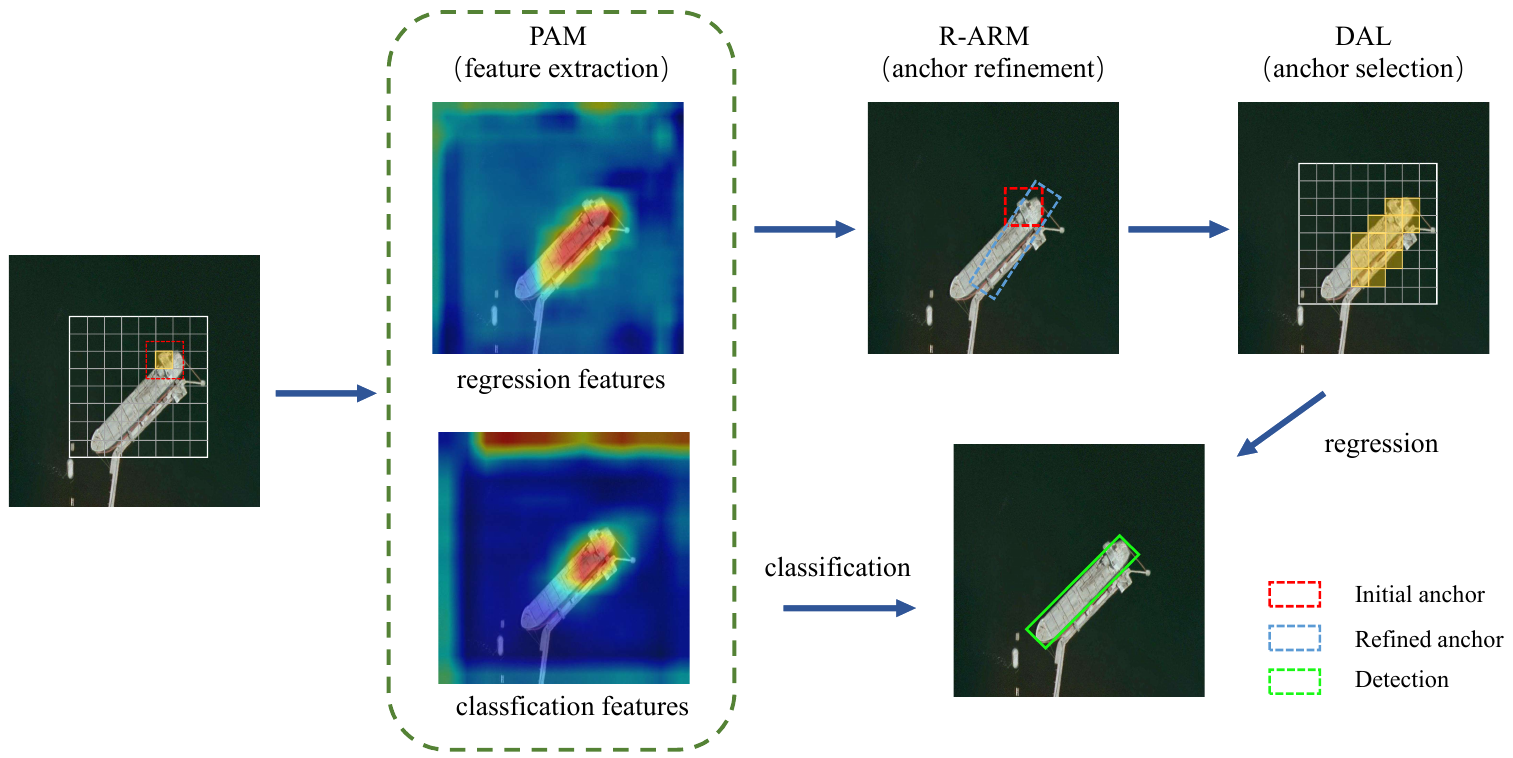} 
	\caption{Illustration of function of the proposed modules in the detection pipeline. The yellow area represents the center of the high-quality anchors}
	\label{Fig5}
\end{figure*}

Finally, the polarization attention weighted features are combined with the original feature pyramid to better extract the critical features. As described in Eq.(\ref{Eq1}), the attention weighted features,  the input features $\mathbf{F}$, and the attention response map $\mathbf{M}$ are merged by element-wise summation to obtain powerful feature representations for accurate object detection. The proposed PAM greatly improves detection performance via optimizing the representation of critical features. The explainable visualization results are shown in Fig.~\ref{Fig5}. It can be seen that PAM can effectively extract the critical features required for different tasks. For example, the extracted regression critical features are evenly distributed on the object, which is helpful to identify the object boundary and accurately localize the target. The classification critical features are concentrated more on the most recognizable part of an object to avoid interference from other parts of the object, and thus the classification results will be more accurate.

\subsection{Rotation Anhcor Refinement Module}
In the existing anchor-based object detectors, classification and regression are performed on densely preset anchors. It is difficult to achieve alignment between anchors and rotation objects owing to the large variation in the scale and orientation of the remote sensing objects. To solve this problem, we proposed a rotation anchor refinement module (R-ARM) to generate high-quality candidates based on critical regression features to reduce the reliance on the priori geometric knowledge of anchors. Given the regression-sensitive feature map extracted by PAM, R-ARM refines the initial anchors to obtain the rotated anchors that better align with the critical regression features. The regions of these high-quality anchors capture the discriminative and semantic features of the object boundary, which helps to achieve accurate localization.

\begin{figure}[t]
	\centering
	\includegraphics[width=0.45\textwidth]{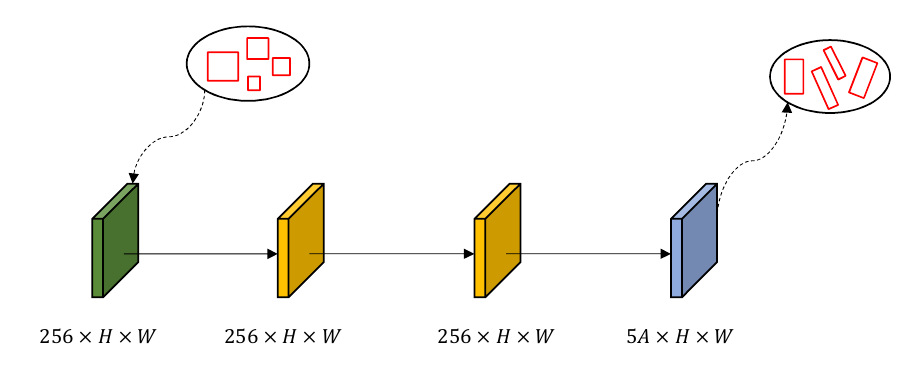} 
	\caption{Illustration of the R-ARM module. $A$ denotes the number of anchors preset at each position of feature map, which is set to 1 in CFC-Net.}
	\label{Fig6}
\end{figure}

The architecture of R-ARM is shown in Fig.~\ref{Fig6}. We preset $A$ initial horizontal anchors at each position of the feature map, which is represented as $(x, y, w, h)$. $(x, y)$ is the center coordinate, and $w, h$ denote the width and height of the anchor, respectively. R-ARM regresses the additional angle $\theta$  and the box offsets of the prior anchor to get the rotation anchor which is expressed as $(x, y, w, h, \theta)$. R-ARM enables anchors to generate refined rotated boxes that are well aligned with the ground-truth objects, and would simultaneously help to capture more critical features for subsequent detection layers. Specifically, we predict offsets $\bm{t^r}$ = $ (t_{x}, t_{y}, t_{w}, t_{h}, t_{\theta})$ for anchor refinement, which are represented as follows:
\begin{equation} 
\begin{aligned}
t_{x}^{r}&=\left(x-x^{a}\right) / w^{a}, \quad t_{y}^{r}=\left(y-y^{a}\right) / h^{a}, \\
t_{w}^{r}&=\log \left(w / w^{a}\right), \quad \quad t_{h}^{r}=\log \left(h / h^{a}\right), \\
t_{\theta}^{r}&=\tan \left(\theta-\theta^{a}\right),
\end{aligned}
\label{Eq6}
\end{equation}
where $x$ and $x^a$ are for the refined box and anchor respectively (likewise for $y, w, h, \theta$).

In CFC-Net, we set $A=1$. It means that only one initial anchor is used, and thus we do not need to carefully set the hyperparameters of angle, aspect ratio, and scale for anchors like the current anchor-based methods, due to the special design of R-ARM after the PAM. Note also that we do not integrate classification prediction in R-ARM as some cascade regression approaches \cite{yang2019r3det, tian2019cascaded}. This is owing to the following considerations: 
\begin{enumerate}
\item [1.] Classification in the refining stage is not accurate enough, and thus it is easy to mistakenly exclude the potential high-quality candidates, resulting in a poor recall of detections.
\item [2.] As mentioned in Section \uppercase\expandafter{\romannumeral1}, there is a gap between classification and regression. The high classification score does not guarantee accurate localization. The training sample selection based on classification confidence in anchor refinement will further degrade the detection performance.
\end{enumerate}

Compared with previous one-stage detectors, CFC-Net needs fewer predefined anchors, but achieves better detection performance with the R-ARM. As illustrated in Fig.~\ref{Fig5}, guided by the critical regression features generated by PAM, the initial square anchor produces a more accurate rotated candidate via R-ARM. The refined anchor aligns well with the high-response region that captures critical features, which provides an effective semantic prior for subsequent localization.

\subsection{Dynamic Anchor Learning}
In the previous sections, we have introduced the critical feature extraction structure and high-quality anchor generation in CFC-Net. However, the misalignment between classification and regression tasks still exists, that is, the high classification scores can not guarantee the accurate localization of the detections. This issue has been widely discussed in many studies \cite{he2019bounding, choi2019gaussian, jiang2018acquisition, feng2018towards}, and some of the work attributed it to the regression uncertainty \cite{choi2019gaussian, feng2018towards}, which reveals that the localization results obtained by the regression are not completely credible. We believe that the gap between classification and regression mainly comes from unreasonable training sample selection \cite{ming2020dynamic}, and further solve this problem from the perspective of critical features. 

Current detectors usually select positive anchors in the label assignment for training according to the IoU between anchors and GT boxes. For simplicity, we denote the IoU between anchors and GT boxes as $IoU_{in}$, while the IoU between the predicted boxes and GT boxes as $IoU_{out}$. The selected positive anchors are supposed to have good semantic information which is conducive to object localization. However, although there is a positive correlation between the classification score and the $IoU_{in}$ (see Fig.~\ref{Fig7}(a)), the high $IoU_{in}$ does not guarantee good localization potential of the anchors, and as shown in Fig.~\ref{Fig7}(b), there is only a weak correlation between the classification confidence and localization capability of predicted boxes. We believe that one of the main causes is that the samples selected according to the $IoU_{in}$  do not align well with the critical features of the objects. 

To resolve the above problems, a Dynamic Anchor Learning (DAL) method is adopted to select samples with strong critical feature capturing ability in the training phase. DAL consists of two parts: dynamic anchor selection (DAS) and matching-sensitive loss (MSL). The rest of this section will elaborate on the implementation of the two strategies.

Firstly, we adopt a new standard called \textit{matching degree} to guide training sample division. It is defined as follows:
\begin{equation} 
m d=\alpha \cdot IoU_{in}+(1-\alpha) \cdot IoU_{out} - u^{\gamma},
\label{Eq7}
\end{equation}
in which $IoU_{in}$ and $IoU_{out}$ are the IoUs between the anchor box and the GT box before and after regression, respectively. $\alpha$ and $\gamma$ are hyperparameters used to weight the influence of different items. $u$ is the penalty term used to suppress the uncertainty during the regression process. The matching degree combines the prior information of spatial alignment, critical feature alignment ability, and regression uncertainty of the anchor to measure its localization capacity. Specifically, for a predefined anchor and its assigned GT box, $IoU_{in}$ is the measure of initial spatial alignment, while $IoU_{out}$ can be used to indicate the critical feature alignment ability. Intuitively, higher $IoU_{out}$ means that the anchor better captures critical regression features and has a stronger localization potential. However, actually, this indicator is unreliable due to the regression uncertainty. It is possible that some high-quality anchors with high $IoU_{in}$ but low $IoU_{out}$ would be mistakenly judged as negative samples\cite{ming2020dynamic}.

Therefore, in Eq.(\ref{Eq7}) we further introduce the penalty term $u$ to alleviate the influence from regression uncertainty. It is defined as follows:

\begin{figure}[t]
	\centering
	\subfloat[]{
		\includegraphics[width=0.22\textwidth]{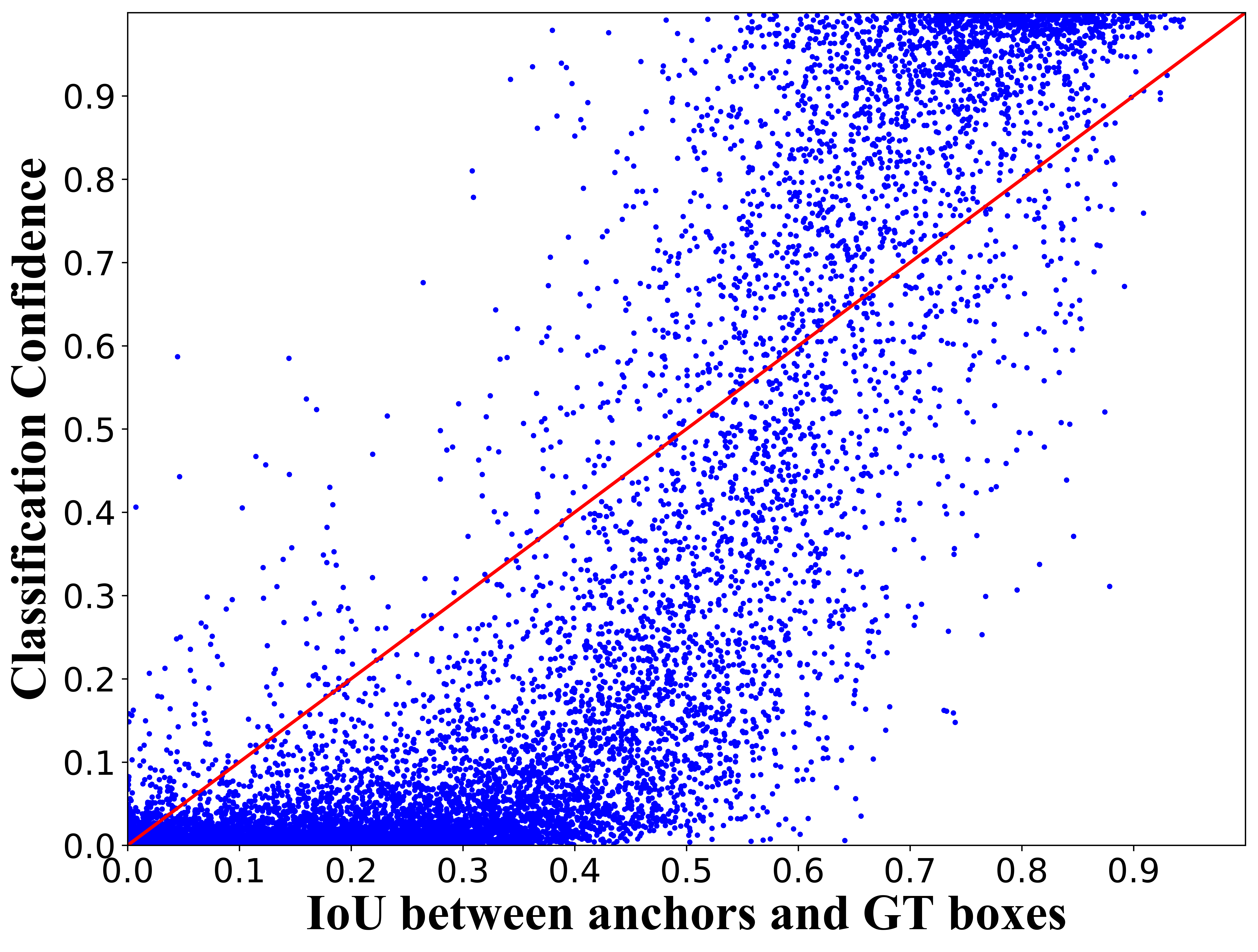}}\hspace{2mm}
	\subfloat[]{
		\includegraphics[width=0.22\textwidth]{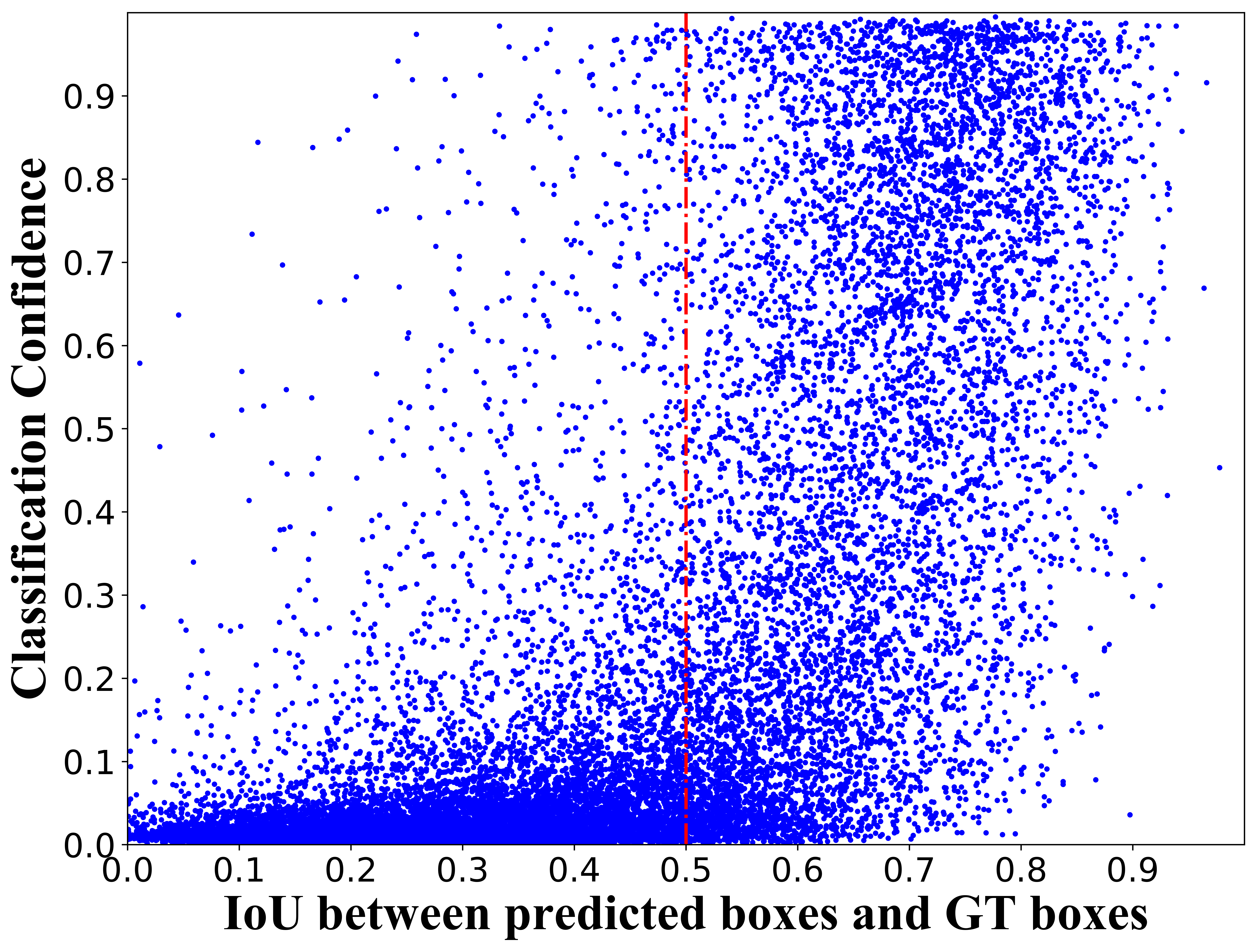}}
	\caption{Analysis of the classification and regression capabilities of anchors that use input IoU for label assignment.}
	\label{Fig7}
\end{figure}

\begin{figure}[t]
	\subfloat[without MSL]{
		\includegraphics[width=0.22\textwidth]{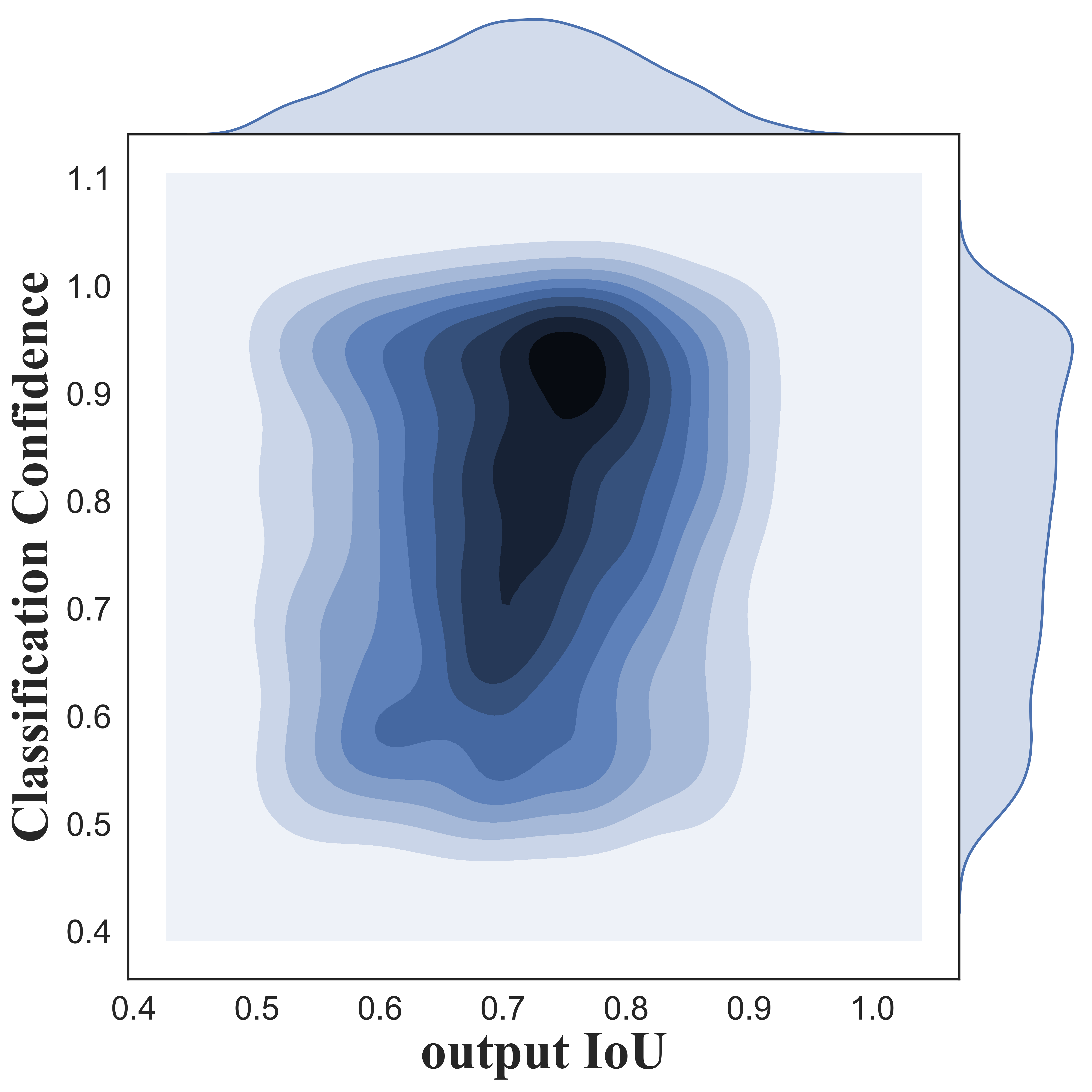}}\hspace{2mm}
	\subfloat[with MSL]{
		\includegraphics[width=0.22\textwidth]{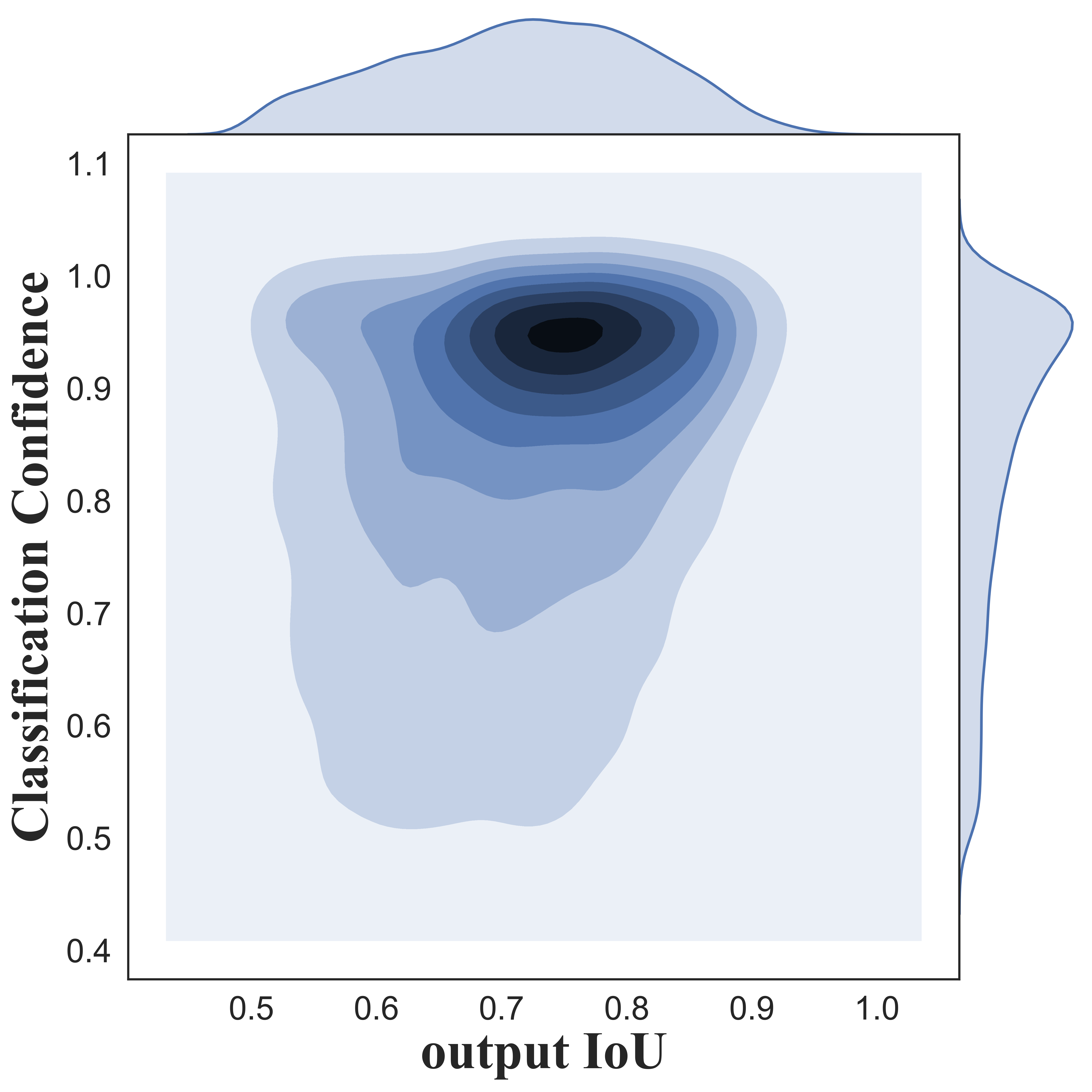}}
	\caption{ The correlation between the output IoU and classification score with and without MSL.}
	\label{Fig8}
\end{figure}

\begin{equation} 
u=|IoU_{in}-IoU_{out}|,
\end{equation}
The change of IoU after regression indicates the probability of incorrect anchor assessment, and we use this to measure regression uncertainty. Uncertainty suppression item $u$ imposes a distrust penalty on samples with excessive IoU change after regression to ensure a reasonable training sample selection. We will confirm in the experimental part that the suppression of uncertainty during regression is the key to take advantage of the critical feature information. 

With the evaluation of the matching degree, we can conduct better training sample selection. We first calculate the matching degree between all anchors and GT boxes in the images, and then candidates with matching degree higher than a certain threshold (set to 0.6 in our experiment) are selected as positive samples, while the rest are negatives. Next, for targets that are not assigned with any positives, the candidate with the highest matching degree will be selected as a positive sample. 

The matching degree measures the ability of feature alignment, and thus the division of positive and negative samples is more reasonable, which would alleviate the misalignment between the classification and regression. It can be seen from Fig.~\ref{Fig5} that DAL dynamically selects anchors that capture the critical regression features for bounding box regression. These high-quality candidates can obtain accurate localization performance after the regression, thereby alleviating the inconsistency before and after the regression, and alleviating the misalignment between classification and regression tasks. 

We further integrate matching degree into the training process to construct a matching-sensitive loss (MSL) to achieve high-performance detection. The classification loss is as follows:
\begin{multline}
L_{cls}=\frac{1}{N_n}\sum_{i \in \psi_n}  FL  \left(p_{i}, p_{i}^{*}\right) \\+ \frac{1}{N_{p}}\sum_{j \in \psi_p} (w_j + 1) \cdot FL\left(p_{j}, p_{j}^{*}\right),
\end{multline}
in which $N_n$ and $N_p$  inidcates the number of all negative and positive anchors, respectively. $\psi_n$ and  $\psi_p$ respectively represent negative and positive samples. $FL(\cdot)$  is focal loss defined as RetinaNet \cite{lin2017focal}. $p^{*}$ is the classification label for anchor ($p^{*}= 1$ if it is positive, while $p^{*}= 0$ otherwise). $w_j$ represents the weighting factor, which is utilized to distinguish positive candidates with different localization ability. For a given target $g$, we first calculate its matching degrees (denoted by $md$) with all preset anchors, among which we then select the matching degrees of positives (denoted by $\bm{md}_{pos}$, and $ \bm{md}_{pos} \subseteq \bm{md} $). Assuming that the maximum value of $\bm{md}_{pos}$ is $md_{max}$, we define a compensation value $\Delta md$ as follows:

\begin{equation} 
\Delta md = 1- md_{max} .
\end{equation}
Subsequently, $\Delta md$  is added to the matching degree of all positive candidates to obtain the weighting factor:

\begin{equation} 
\bm{w} =  \bm{md}_{pos}+\Delta md .
\end{equation}
The weighting factor improves the contribution of the positive samples to the loss during the training process. In this way, the classification branch can discriminate anchors with different capabilities to capture critical features. Compared with the commonly used method that treats all positive anchors equally, this discriminative approach helps to distinguish positive samples of different localization ability. By involving the localization information of anchors into the classification loss, the classifier can output more reliable classification confidence to select the detections with good localization, thereby bridging the gap between classification and regression.

Since matching degree measures the localization ability of anchors, it can be further used to promote high-quality localization. The matching-sensitive regression loss is defined as follows:

\begin{equation} 
L_{reg}=\frac{1}{N_p}\sum_{j \in \psi_p} w_j \cdot L_{smooth_{L_1}}\left(\bm{t_{j}}, \bm{t_{j}^{*}}\right) ,
\end{equation}
where $L_{smooth_{L_1}}$ represents the smooth-$L_1$ loss\cite{girshick2015fast}. $\bm{t}$ and $\bm{t^{*}}$ are offsets for the predicted boxes and target boxes, respectively. The weighted regression loss can adaptively pay more attention to the samples with high localization potential rather than good initial spatial alignment, and thus better detection performance would be achieved after the training. It can be seen from Fig.~\ref{Fig8}(a) that the detectors trained with normal smooth-$L_1$ loss exhibits a weak correlation between the classification score and localization ability of the detections, which causes the predictions selected by the classification confidence to be unreliable. After training with a matching-sensitive loss function, as shown in Fig.~\ref{Fig8}(b), the detection with better localization performance will also achieve higher classification confidence, facilitating the selection of high-quality detection based on the classification score. The above analysis confirms the effectiveness of the matching-sensitive loss.

Dynamic anchor selection strategy and matching-sensitive loss can also be employed to the anchor refinement stage, and thus the multitask loss for CFC-Net is defined as follows: 
\begin{equation} 
L=L_{cls}(p, p^{*}) + \lambda_{1}L_{ref}\left(\boldsymbol{t^r}, \boldsymbol{t}^{*}\right) + \lambda_{2}L_{reg}\left(\boldsymbol{t}, \boldsymbol{t}^{*}\right),
\end{equation}
where $L_{cls}(p, p^{*})$, $L_{ref}\left(\boldsymbol{t^r}, \boldsymbol{t}^{*}\right) $, and $L_{reg}\left(\boldsymbol{t}, \boldsymbol{t}^{*}\right)$ are the classification loss, anchor refinement loss, and regression loss, respectively. $\bm{t^r}$, $\bm{t}$ denotes the predicted offsets of refined anchors and detection boxes, respectively. $\bm{{t}^{*}}$ represents the offsets of GT boxes. These loss items are balanced via parameters $\lambda_{1}$ and $\lambda_{2}$ (we set $\lambda_{1}=\lambda_{2}=0.5$ in our experiments).

\section{Experiments}

\subsection{Datasets}
Experiments are conducted on three public remote sensing datasets: HRSC2016, DOTA, and UCAS-AOD. The ground-truth boxes in these datasets are annotated with oriented bounding box.  

HRSC2016 \cite{liu2017high} is a high resolution remote sensing ship detection dataset with a total of 1061 images. The image sizes range from 300$\times$300 to 1500$\times$900. The entire dataset is divided into training set, validation set, and test set, including 436, 181, and 444 images, respectively. The images are resized to two scales of 416$\times$416 and 800$\times$800 in our experiments.

DOTA \cite{xia2018dota}  is the largest publicly available dataset for oriented object detection in remote sensing images. DOTA includes 2806 aerial images with 188,282 annotated instances. There are 15 categories in total, including  plane (PL), baseball diamond (BD), bridge (BR), ground track field (GTF), small vehicle (SV), large vehicle (LV), ship (SH), tennis court (TC), basketball court (BC), storage tank (ST), soccer ball field (SBF), roundabout (RA), harbor (HA), swimming pool (SP) and helicopter (HC). Note that images in DOTA are too large, we crop the original images into 800$\times$800 patches with the stride 200 for training and testing. 

UCAS-AOD \cite{zhu2015orientation} is an aerial aircraft and car detection dataset, which contains 1510 images collected from Google Earth. It includes 1000 planes images and 510 cars images in total. Since there is no official division of this dataset. we randomly divide it into training set, validation set, and test set as 5:2:3. All images in UCAS-AOD are resized to 800$\times$800 in the experiments.

\subsection{Implementation Details}
The backbone of our CFC-Net is ResNet-50 \cite{he2016deep}. The model is pre-trained on the ImageNet and fine-tuned on remote sensing image datasets. We utilize the feature pyramid of $P_3, P_4, P_5,P_6,P_7$ to detect multi-scale objects. For each position of the feature map, only one anchor is set to regress the nearby objects. We use random flipping, rotation, and HSV jittering for data augmentation. We take matching degree threshold of positives to be 0.4 for the refinement stage, while 0.6 for detection stage for high-quality detections. 

The mean Average Precision (mAP) defined in PASCAL VOC object detection challenge \cite{everingham2010pascal} is used as the evaluation metric for all experiments. For a fair comparison with other methods, HRSC2016 dataset and UCAS-AOD dataset use the mAP metric defined in PASCAL VOC 2007 challenge, while DOTA adopts PASCAL VOC 2012 definition. Our ablation studies are conducted on the HRSC2016 dataset since remote sensing ships often have a large aspect ratio and scale variation, which are major challenges for object detection in optical remote sensing images. In the ablation studies, all images are scaled to 416$\times$416 without data augmentation.

We train the model with the batch size set to 8 on RTX 2080Ti GPU.  The network is trained with Adam optimizer. The learning rate is set to 1e-4 and  is divided by 10 at each decay step. The total iterations of HRSC2016, UCAS-AOD, and DOTA are 10k, 5k, and 40k, respectively.

\begin{table}[t]
	\renewcommand\arraystretch{1.1}
	\centering
	\caption{Effects of each component of CFC-Net.}
	\setlength{\tabcolsep}{2.5mm}{
	\begin{tabular}{{c|ccccc}}
		\toprule
		& \multicolumn{5}{c}{Different Variants } \\ 
		\midrule
		with PAM?      & \texttimes   &\checkmark & \texttimes   &\checkmark  &\checkmark \\ 
		with DAL?      & \texttimes   & \texttimes   &\checkmark  &\checkmark &\checkmark \\ 
		with R-ARM? & \texttimes   & \texttimes    & \texttimes    & \texttimes    &\checkmark \\ 
		\midrule
		mAP              &     70.5     &    76.2      &    78.7     &  83.8&  $\textbf{86.3}$      \\ 
		\bottomrule
	\end{tabular}}

	\label{Table1}
\end{table}

\begin{table}[t]
	\renewcommand\arraystretch{1.1}
	\centering
	\caption{Ablation study of the proposed PAM.}
	\setlength{\tabcolsep}{2.5mm}{
	\begin{tabular}{{c|cccc}}
		\toprule
		& \multicolumn{4}{c}{Different Variants } \\ 
		\midrule
		+ dual FPN      			& \texttimes   &\checkmark  & \checkmark         &\checkmark \\ 
		+ attention     			& \texttimes   & \texttimes    &\checkmark         &\checkmark \\ 
		+ polarization function  & \texttimes   & \texttimes    & \texttimes      &\checkmark \\ 
		\midrule
		mAP              &     70.5     &    72.1     &     74.9   &  $\textbf{76.2}$      \\ 
		\bottomrule
	\end{tabular}}
	\label{Table2}
\end{table}

\begin{figure}[t]
	\centering
	\includegraphics[width=0.47\textwidth]{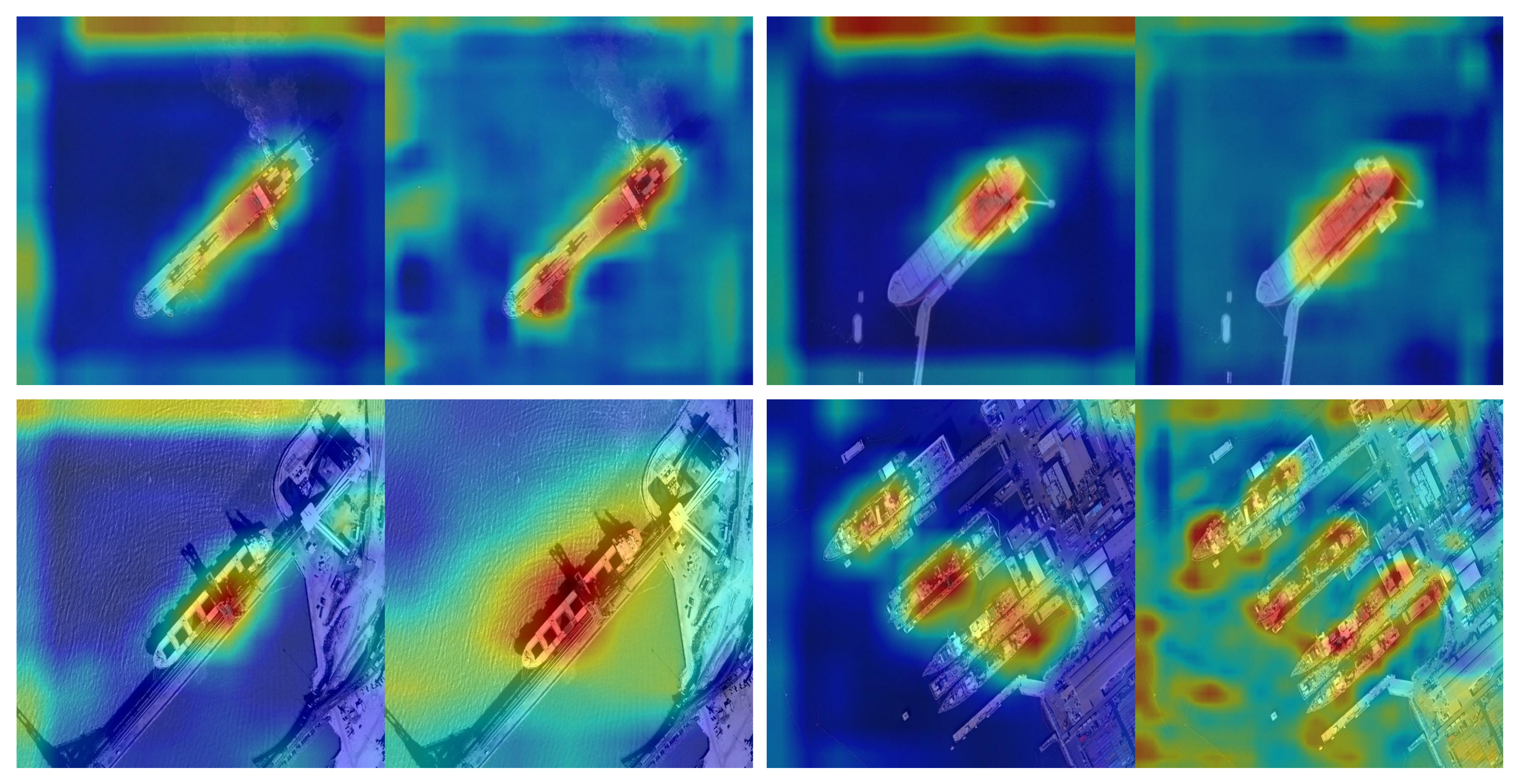} 
	\caption{Visualization results of critical features for classification and regression tasks. The left side of each pair of images shows the distribution of critical features for classification task, while the right side for regression task.}
	\label{Fig9}
\end{figure}

\subsection{Ablation Study}

\subsubsection{Evaluation of different components}
We conduct a component-wise experiment on HRSC2016 dataset to verify the contribution of the proposed components. The experimental results are shown in Table \ref{Table1}. Since only one anchor is preset, it is difficult to capture the critical features required to identify the object, so the baseline model only achieves the mAP of 70.5\%. Using the PAM, the detection performance is increased by 5.7\%, indicating that the PAM module effectively constructs more powerful feature representations, so that even one preset anchor can make good use of critical features to achieve accurate detection. The performance of the model is improved by 8.2\% using DAL, due to its ability of selecting high-quality anchors with good critical feature alignment in the learning process. The simultaneous use of DAL and PAM achieves a mAP of 83.8\%, indicating that the two methods do not conflict and can effectively improve the detection performance. The proposed  R-ARM refines the horizontal anchors to obtain high-quality rotated anchors. It further improves the performance by 2.5\%. Finally, CFC-Net reaches the mAP of 86.3\% with an increase of 15.8\% compared with the naive model, proving the effectiveness of our framework.

\subsubsection{Evaluation of PAM}

To verify the effect of the proposed PAM, we conduct some comparative experiments on HRSC2016 dataset. The results are shown in Table \ref{Table2}. By using dual FPN to extract independent features for classification and regression branches, the detection performance is improved by 1.6\% compared with the baseline model. Although dual FPN seperates features for different tasks and  improves performance, the features are not fully utilized. When we adopt the attention mechanism based on dual FPN, a further improvement of 2.8\%  is achieved. It indicates that the attention mechanism enables the features of different branches to better respond to the discriminative parts of the objects. Through the processing of the polarization function, the discriminative parts of the critical classification features are strengthened, while the high response regions in the critical regression features are suppressed to find more clues to further improve localization results. The improvement of 1.3\% based on the attention-based model confirms our viewpoint. These experiments prove that the proposed components of PAM can effectively improve the detection performance. 

Some visualization results are shown in Fig.~\ref{Fig9}. It can be seen that the heatmap induced by PAM accurately responds to the area of task-sensitive critical features. The discriminative areas required for classification are often concentrated in the local part of objects, such as the stern and bow of ships. Meanwhile, the clues required for regression are more likely to be distributed on the edge of the objects or the contextual information.

\begin{table}
	\renewcommand\arraystretch{1.1}
	\centering
	\caption{Ablation study of DAL.}
	\setlength{\tabcolsep}{2.5mm}{
	\begin{tabular}{{c|cccc}}
		\toprule
		& \multicolumn{4}{c}{Different Variants} \\ 
		\midrule
		with Input IoU? 			    &\checkmark &\checkmark & \checkmark &\checkmark \\ 
		with Output IoU?		    & \texttimes   &\checkmark &\checkmark  &\checkmark\\ 
		Uncertainty Supression?    & \texttimes   &\texttimes	&\checkmark  &\checkmark\\ 
		Matching Sensitive Loss?  & \texttimes   &\texttimes 	& \texttimes    &\checkmark\\ 
		\midrule
		mAP&     70.5     &    71.3      &    76.2     &   $\textbf{78.7}$      \\ 
		\bottomrule
	\end{tabular}}
	\label{Table3}
\end{table}

\begin{table}[t]
	\renewcommand\arraystretch{1.1}
	\centering
	\caption{ Ablation study of the proposed R-ARM.}
	\setlength{\tabcolsep}{4.2mm}{
	\begin{tabular}{c|ccc}
		\toprule
		refinement stages & 0 & 1 & 2  \\ 
		\midrule
		mAP  &  83.8 &  $\textbf{86.3}$  & 84.5 \\ 
		\bottomrule
	\end{tabular}}
	\label{Table4}
\end{table}

\begin{table}[t]
	\renewcommand\arraystretch{1.1}
	\centering
	\caption{Analysis of influence of different hyperparameters.}
	\begin{tabular}{c|cc|c|cc|c|cc}
		\toprule
		$\gamma$     & $\alpha$   & $\text{mAP}$& $\gamma$   & $\alpha$   & $\text{mAP}$ & 	$\gamma$   &  $\alpha$& $\text{mAP}$   \\
		\midrule
		\multirow{5}{*}{2} & 0.2 &76.2& \multirow{5}{*}{3} &0.2 &69.8 & \multirow{5}{*}{4} & 0.2 & 47.4 \\
								 & 0.3 & 72.7 &                    		   & 0.3 & 78.5 &                    			& 0.3 & 76.9\\ 
								 & 0.5 & 70.0 &                    		   & 0.5 & 74.4 &                    			& 0.5 & $\textbf{78.7}$ \\ 
								 & 0.7 & 71.7 &                   		   & 0.7 & 69.4 &                    			& 0.7 & 77.3 \\ 
								 & 0.9 & 43.9 &                   		   & 0.9 & 69.1 &                    			& 0.9 & 72.1 \\
\bottomrule
	\end{tabular}
\label{Table5}
\end{table}

\begin{table}[t]
	\renewcommand\arraystretch{1.2}
	\centering
	\caption{Comparisons with different methods on HRSC2016 dataset.}
	\setlength{\tabcolsep}{2.3mm}{
	\begin{tabular}{c|ccc|c}
		\toprule
		Methods & Backbone & Size & NA & mAP \\
		\midrule
		\multicolumn{1}{l}{\textit{Two-stage:}}&\multicolumn{4}{l}{}  \\
		\midrule
		$\text{R}^2\text{CNN}$\cite{jiang2017r2cnn}        & ResNet101   &800$\times$800     & 21   & 73.1 \\
		RC1\&RC2\cite{liu2017high}         &  VGG16      &  -                &  -   &  75.7 \\
		RRPN\cite{ma2018arbitrary}        &  ResNet101  & 800$\times$800    &  54  & 79.1  \\
		$\text{R}^2\text{PN}$\cite{zhang2018toward}        &  VGG16      &  -                &   24 &  79.6 \\
		RoI Trans. \cite{ding2019learning}&  ResNet101  & 512$\times$800    &   5  &   86.2\\
		Gliding Vertex\cite{xu2020gliding}&  ResNet101  & 512$\times$800    &   5  &  88.2 \\
		\midrule
		\multicolumn{1}{l}{\textit{Single-stage:}}&\multicolumn{4}{l}{}  \\
		\midrule
		RRD\cite{liao2018rotation}        &  VGG16      & 384$\times$384    & 13   &  84.3 \\
		$\text{R}^3\text{Det}$\cite{yang2019r3det}         &  ResNet101  & 800$\times$800    &   21 &  89.3 \\
		R-RetinaNet\cite{lin2017focal}                       &  ResNet101  & 800$\times$800    &  121 &  89.2 \\
		CFC-Net                        &  ResNet50    & 416$\times$416    & 1    &  86.3 \\
		CFC-Net (aug)                       &  ResNet50    & 800$\times$800    &1   &  88.6 \\
		CFC-Net (aug)                        &  ResNet101    & 800$\times$800    & 1    & 89.5   \\
		CFC-Net (aug + ms)                        &  ResNet101  & 800$\times$800    & 1    &   $\textbf{89.7}$\\
		\bottomrule
	\end{tabular}}
	\label{Table6}
\end{table}

\begin{figure}[t]
	\centering
	\includegraphics[width=0.45\textwidth]{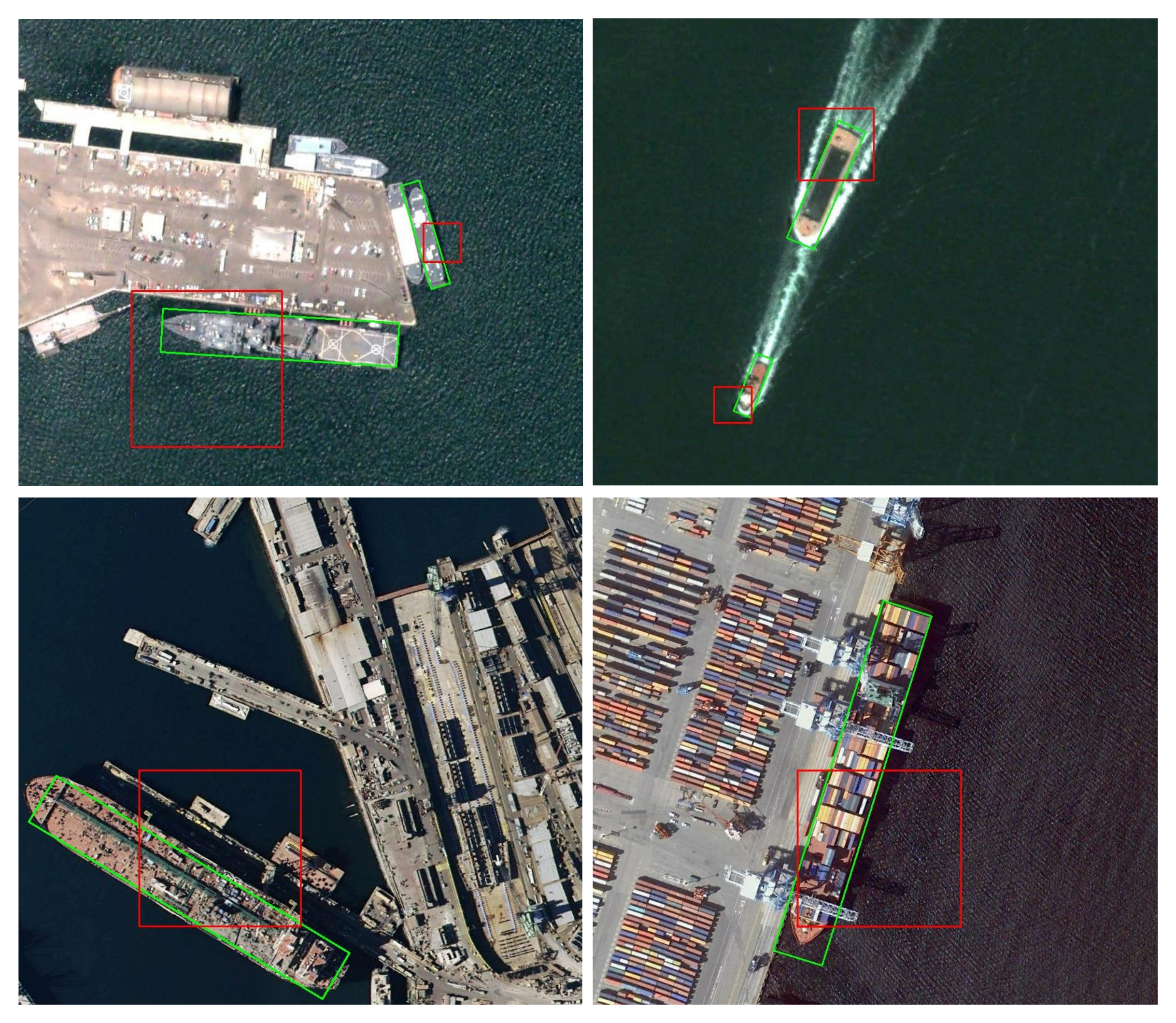} 
	\caption{Detection results on HRSC2016 dataset with our method. The red boxes and green boxes indicate the anchor boxes and detection results respectively.}
	\label{Fig10}
\end{figure}

\subsubsection{Evaluation of DAL}
We conduct component-wise experiments to verify the contribution of the DAL. The experimental results are shown in Table \ref{Table3}, in which input IoU, output IoU and regression uncertainty are denoted by the three terms in Eq.(\ref{Eq7}), respectively. For the variants with output IoU,  $\alpha$ is set to 0.8 for stable training, and the detection performance slightly increases from 70.5\% to 71.3\%. It indicates that using output IoU alone is insignificant for training sample selection. With the suppression of regression uncertainty, the prior space alignment and posterior critical feature alignment can work together to dramatically improve the performance by 5.7\% compared with the baseline. Furthermore, matching degree guided loss function effectively distinguishes anchors with differential localization capability, and thus model using the matching sensitivity loss function achieves the mAP of 78.7\%,

\subsubsection{Evaluation of R-ARM}
Based on DAL and PAM, we further conduct experiments to verify the effect of the proposed R-ARM and explore the influence of the number of refinement stages. For the model without R-ARM, we set the matching degree threshold of positives to 0.4. For the one-stage refinement model, the thresholds of the refinement stage and the detection stage are set to 0.4 and 0.6, respectively.  The thresholds are set to 0.4, 0.6, and 0.8 for two-stage refinement module. As shown in Table \ref{Table4}, with one-stage R-ARM, the performance is increased by 2.5\%. It can be attributed to the fact that the refined proposals learned from horizontal anchors provide high-quality samples, and these candidates are better aligned with critical features of objects. However, adopting two-stage R-ARM drops the performance by 1.8\% compared with the one-stage R-ARM. It may be that as the threshold increases in detection stage, the number of positives that higher than the current matching degree threshold decreases sharply, leading to insufficient positive samples and a serious imbalance of positives and negatives. Thus we use one stage R-ARM in CFC-Net.

\begin{figure*}[t]
	\centering
	\includegraphics[width=0.95\textwidth]{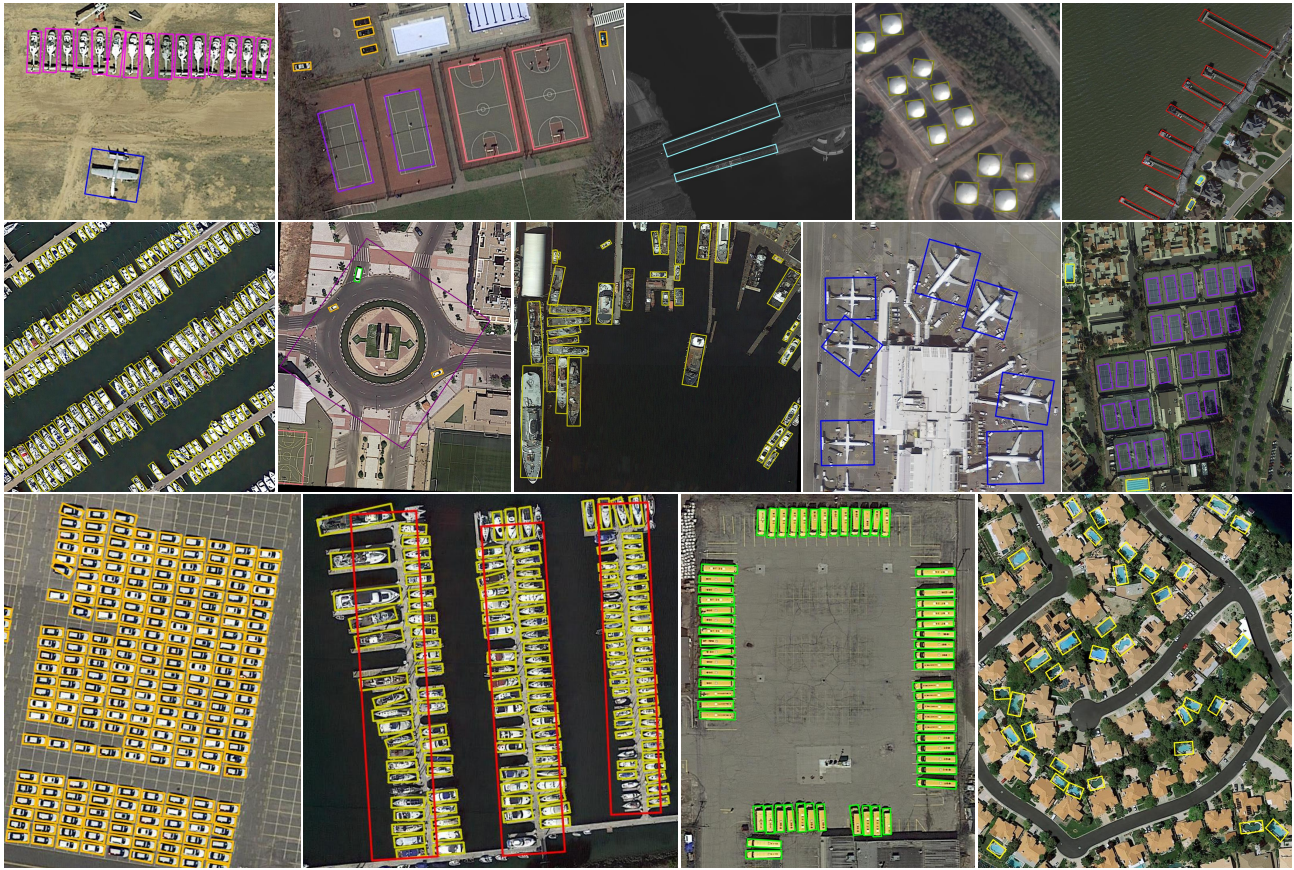} 
	\caption{Visualization of detection results on DOTA dataset with our method.}
	\label{Fig11}
\end{figure*}

\begin{table*}[t]
	\renewcommand\arraystretch{1.2}
	\centering
	\caption{  Performance evaluation of OBB task on DOTA dataset.}
	\setlength{\tabcolsep}{1.5mm}
	{
	\begin{tabular}{c|ccccccccccccccc|c}
		\toprule
		Methods   & PL & BD & BR & GTF & SV & LV & SH & TC & BC & ST & SBF & RA & HA & SP & HC & mAP \\ 
		\midrule
		FR-O\cite{xia2018dota}                         & 79.09  &69.12  &17.17  &63.49   &34.20  &37.16  &36.20  &89.19  &69.60  &58.96  &49.40  &52.52  &46.69  &44.80  &46.30  &52.93\\ 
		R-DFPN\cite{yang2018automatic}                  & 80.92  &65.82  &33.77  &58.94   &55.77  &50.94  &54.78  &90.33  &66.34  &68.66  &48.73  &51.76  &55.10  &51.32  &35.88  &57.94 \\
		$\text{R}^2\text{CNN}$\cite{jiang2017r2cnn}     & 80.94  &65.67  &35.34  &67.44   &59.92  &50.91  &55.81  &90.67  &66.92  &72.39  &55.06  &52.23  &55.14  &53.35  &48.22  &60.67 \\ 
		RRPN\cite{ma2018arbitrary}	                    & 88.52  &71.20  &31.66  &59.30   &51.85  &56.19  &57.25  &90.81  &72.84  &67.38  &56.69  &52.84  &53.08  &51.94  &53.58  &61.01\\ 
		ICN\cite{azimi2018towards} 	                   & 81.36  &74.30  &47.70  &70.32   &64.89  &67.82  &69.98  &90.76  &79.06  &78.20  &53.64  &62.90  &67.02  &64.17  &50.23  &68.16\\
		RoI Trans.\cite{ding2019learning}                 & 88.64  &78.52  &43.44  &$\textbf{75.92}$   &68.81  &73.68  &83.59  &90.74  &77.27  &81.46  &58.39  &53.54  &62.83  &58.93  &47.67  &69.56\\
		CAD-Net\cite{zhang2019cad}                      & 87.80  &$\textbf{82.40}$  &49.40  &73.50   &71.10  &63.50  &76.70  &$\textbf{90.90}$  &79.20  &73.30  &48.40  &60.90  &62.00  &67.00  &62.20  &69.90 \\ 
		DRN\cite{pan2020dynamic}                      & 88.91  &80.22  &43.52  &63.35   &73.48  &70.69  &84.94  &90.14  &83.85  &84.11  &50.12  &58.41  &67.62  &$\textbf{68.60 }$  &52.50  &70.70  \\ 
		$\text{O}^2\text{-DNet}$\cite{wei2020oriented}   & 89.31  &82.14  &47.33  &61.21   &71.32  &74.03  &78.62  &90.76  &82.23  &81.36  &60.93  &60.17  &58.21  &66.98  &61.03  &71.04\\
		$\text{R}^3\text{Det}$\cite{yang2019r3det}   	    &89.54  &81.99 &48.46  &62.52   &70.48  &74.29  &77.54  &90.80  &81.39  &83.54  &61.97  &59.82  &65.44  &67.46  &60.05  &71.69 \\ 
		SCRDet\cite{yang2019scrdet}		                & $\textbf{89.98}$  &80.65  &52.09  &68.36   &68.36  &60.32  &72.41  &90.85  &$\textbf{87.94}$  &$\textbf{86.86}$  &$\textbf{65.02}$  &$\textbf{66.68}$  &66.25  &68.24  &$\textbf{65.21}$  &72.61\\ 
		CFC-Net (ours)  												       &89.08  &80.41  &$\textbf{52.41}$   &70.02  &$\textbf{76.28}$  &$\textbf{78.11}$  &$\textbf{87.21}$  &90.89  &84.47  &85.64  &60.51  &61.52  &$\textbf{67.82}$ &68.02  &50.09 &$\textbf{73.50}$ \\  
		\bottomrule
	\end{tabular}}
\label{Table7}
\end{table*}  

\subsubsection{Hyper-parameters}
In order to find suitable hyperparameter settings, we conduct parameter sensitivity experiments, and the results are shown in Table \ref{Table5}. As the $\alpha$ is  reduced appropriately, the effect of feature alignment increases, and the mAP increases. For example, on condition that $\gamma$ is equal to 4, as $\alpha$ decreases from 0.9 to 0.5, the mAP increases from 72.1\% to 78.7\%. It indicates that under the premise of uncertainty suppression, the feature alignment represented by the $IoU_{out}$ is conducive  to selecting anchors with high localization capabilities. However, when $\alpha$ is extremely small, the performance drops sharply (like $\gamma=4$), because the anchors selected by the dominant output IoU may contain too many false-positive samples. In this case, prior space alignment can help alleviate this problem and make anchor selection more stable. In addition, as $\gamma$ decreases, the ability to suppress disturbance samples is stronger, but it may also suppress the mining of potential positives, resulting in performance degradation.

\subsection{Main Results and Analysis}
\subsubsection{Results on HRSC2016}
HRSC2016 contains lots of remote sensing oriented ships with a large aspect ratio, scales and arbitrary orientations. Our method achieves competitive performances on HRSC2016 dataset. As shown in Table \ref{Table6}, ‘aug’ represents using data augmentation, ‘ms’ denotes multi-scale training and testing, and NA is the number of preset anchors at each location of feature maps. The proposed CFC-Net achieves the mAP of 86.3\% when input images are rescaled to 416$\times$416 without data augmentation, which is comparable to many previous advanced methods. With data augmentation and the input image resized to 800$\times$800, our method reaches the mAP of 88.6\%, which is better than many recent methods. Further using multi-scale training and testing, our method achieves state-of-the-art performance on HRSC2016 dataset among the compared  methods, reaching the mAP of 89.7\%. 

It is worth mentioning that our approach uses only one horizontal anchor at each position of feature maps, but outperforms the frameworks with a large number of anchors. These results show that it is unnecessary to preset a large number of rotated anchors for oriented object detection. Instead, the more important thing is to select high-quality anchors and capture the critical features for object recognization. For instance,  the anchors in Fig.~\ref{Fig10} have low IoUs with targets in the images and will be regarded as negatives in most detectors. But they actually have a strong potential for accurate localization. CFC-Net effectively utilizes these anchors to achieve efficient and accurate prediction. Note that our model is a single-stage detector, and the feature maps used are $P_3-P_7$. Compared with the $P_2-P_6$ for two-stage detectors, the total amount of positions that need to set anchor is fewer, so the inference speed is faster. With the input image resized to 800$\times$800, our model reaches 28 FPS on RTX 2080 Ti GPU.

\begin{table}[t]
	\renewcommand\arraystretch{1.2}
	\centering	
	\caption{Detection results on UCAS-AOD dataset.}
	\setlength{\tabcolsep}{4.2mm}{
		\begin{tabular}{c|cc|cc}
			\toprule
			Methods & car  & airplane & mAP \\ 
			\midrule
			YOLOv3\cite{redmon2018yolov3}                   		& 74.63 	 & 89.52 				& 82.08 \\ 
			R-RetinaNet\cite{lin2017focal}                               				& 84.64     &  $\textbf{90.51}$ & 87.57    \\ 
			FR-O\cite{xia2018dota}                     			& 86.87      & 89.86     			 & 88.36 \\ 
			RoI Trans. \cite{ding2019learning}		& 88.02 	& 90.02 				&89.02     \\
			CFC-Net                          						 & $\textbf{89.29}$ 		& 88.69 &  $\textbf{89.49}$   \\ 
			\bottomrule
	\end{tabular}}
	\label{Table8}
\end{table}

\subsubsection{Results on DOTA}
We compare the proposed approach with other state-of-the-art methods on DOTA dataset. As shown in Table \ref{Table8}, we achieve the mAP of 73.50\%, which reaches the best performance among the compared  methods. Some detection results on DOTA are shown in Fig.~\ref{Fig11}. It can be seen from the illustration that even though only one anchor is used, our CFC-Net still accurately detects densely arranged small objects (such as ships, small vehicles, and large vehicles in the third row). In addition, the proposed detector also adapts well to the scale variations and accurately locates objects of different scales. Take the second one (from the left) in the second row for example, the precise detections of both large-scale roundabout and small vehicles at different scales are achieved through the feature pyramid with only one prior anchor at each location. Besides, as shown in the third figure and the fifth figure in the first row, our method can use a few square anchors to detect objects with very large aspect ratios (such as bridges and harbors here), These detections indicate that it is not essential for preset anchors to have a good spatial alignment with the objects, while the key is to effectively identify and capture the critical features of the objects. The utilized matching degree measures the critical feature capturing ability of anchors, and on this basis, the DAL strategy performs a more reasonable selection of training samples to achieve high-quality detection.

\subsubsection{Results on UCAS-AOD}
Experimental results in Table \ref{Table7} show that our CFC-Net achieves the best performance among the compared detectors, reaching the mAP of 89.49\%. Note that the original YOLOv3\cite{redmon2018yolov3} and RetinaNet\cite{lin2017focal} are proposed for generic object detection, and the objects are annotated with horizontal bounding box. To make a fair comparison, we introduce an additional angle dimension and perform angle prediction to achieve rotation object detection. Our method outperforms the other compared single-stage detectors, and even better than some advanced two-stage detectors. Besides, the detection performance of small vehicles is excellent, which indicates that our method is robust to densely arranged small objects.

\section{Conclusion}
In this article, we introduce the concept of critical features and prove its importance for high-performance object detection through experiments and observations. On this basis, a Critical Feature Capturing network (CFC-Net) is proposed to optimize the one-stage detector from three aspects: feature representation, anchor refinement, and training sample selection. Specifically, decoupled classification and regression critical features are extracted through the polarization attention mechanism module based on dual FPN. Next, the rotation anchor refinement is performed on one preset anchor to obtain high-quality rotation anchors, which can be better aligned with critical features. Finally, matching degree is adopted to measure the ability of anchors to capture critical features, so as to select positive candidates with high localization potential. As a result, the inconsistency between classification and regression is alleviated and high-quality detection performance can be achieved. Extensive experiments on three remote sensing datasets verify the effectiveness of the proposed method.

\ifCLASSOPTIONcaptionsoff
  \newpage
\fi



\bibliographystyle{IEEEtran}
\bibliography{refer.bib}

\begin{thebibliography}{10}
\providecommand{\url}[1]{#1}
\csname url@samestyle\endcsname
\providecommand{\newblock}{\relax}
\providecommand{\bibinfo}[2]{#2}
\providecommand{\BIBentrySTDinterwordspacing}{\spaceskip=0pt\relax}
\providecommand{\BIBentryALTinterwordstretchfactor}{4}
\providecommand{\BIBentryALTinterwordspacing}{\spaceskip=\fontdimen2\font plus
\BIBentryALTinterwordstretchfactor\fontdimen3\font minus
  \fontdimen4\font\relax}
\providecommand{\BIBforeignlanguage}[2]{{%
\expandafter\ifx\csname l@#1\endcsname\relax
\typeout{** WARNING: IEEEtran.bst: No hyphenation pattern has been}%
\typeout{** loaded for the language `#1'. Using the pattern for}%
\typeout{** the default language instead.}%
\else
\language=\csname l@#1\endcsname
\fi
#2}}
\providecommand{\BIBdecl}{\relax}
\BIBdecl

\bibitem{li2012automatic}
Y.~Li, X.~Sun, H.~Wang, H.~Sun, and X.~Li, ``Automatic target detection in
  high-resolution remote sensing images using a contour-based spatial model,''
  \emph{IEEE Geoscience and Remote Sensing Letters}, vol.~9, no.~5, pp.
  886--890, 2012.

\bibitem{han2014efficient}
J.~Han, P.~Zhou, D.~Zhang, G.~Cheng, L.~Guo, Z.~Liu, S.~Bu, and J.~Wu,
  ``Efficient, simultaneous detection of multi-class geospatial targets based
  on visual saliency modeling and discriminative learning of sparse coding,''
  \emph{ISPRS Journal of Photogrammetry and Remote Sensing}, vol.~89, pp.
  37--48, 2014.

\bibitem{han2014object}
J.~Han, D.~Zhang, G.~Cheng, L.~Guo, and J.~Ren, ``Object detection in optical
  remote sensing images based on weakly supervised learning and high-level
  feature learning,'' \emph{IEEE Transactions on Geoscience and Remote
  Sensing}, vol.~53, no.~6, pp. 3325--3337, 2014.

\bibitem{zhu2010novel}
C.~Zhu, H.~Zhou, R.~Wang, and J.~Guo, ``A novel hierarchical method of ship
  detection from spaceborne optical image based on shape and texture
  features,'' \emph{IEEE Transactions on geoscience and remote sensing},
  vol.~48, no.~9, pp. 3446--3456, 2010.

\bibitem{eikvil2009classification}
L.~Eikvil, L.~Aurdal, and H.~Koren, ``Classification-based vehicle detection in
  high-resolution satellite images,'' \emph{ISPRS Journal of Photogrammetry and
  Remote Sensing}, vol.~64, no.~1, pp. 65--72, 2009.

\bibitem{yang2019road}
X.~Yang, X.~Li, Y.~Ye, R.~Y. Lau, X.~Zhang, and X.~Huang, ``Road detection and
  centerline extraction via deep recurrent convolutional neural network
  {UU-Net},'' \emph{IEEE Transactions on Geoscience and Remote Sensing},
  vol.~57, no.~9, pp. 7209--7220, 2019.

\bibitem{ji2019vehicle}
H.~Ji, Z.~Gao, T.~Mei, and B.~Ramesh, ``Vehicle detection in remote sensing
  images leveraging on simultaneous super-resolution,'' \emph{IEEE Geoscience
  and Remote Sensing Letters}, vol.~17, no.~4, pp. 676--680, 2019.

\bibitem{liu2019multi}
N.~Liu, Z.~Cao, Z.~Cui, Y.~Pi, and S.~Dang, ``Multi-layer abstraction saliency
  for airport detection in {SAR} images,'' \emph{IEEE Transactions on
  Geoscience and Remote Sensing}, vol.~57, no.~12, pp. 9820--9831, 2019.

\bibitem{wu2018inshore}
F.~Wu, Z.~Zhou, B.~Wang, and J.~Ma, ``Inshore ship detection based on
  convolutional neural network in optical satellite images,'' \emph{IEEE
  Journal of Selected Topics in Applied Earth Observations and Remote Sensing},
  vol.~11, no.~11, pp. 4005--4015, 2018.

\bibitem{li2018hsf}
Q.~Li, L.~Mou, Q.~Liu, Y.~Wang, and X.~X. Zhu, ``{HSF-Net}: {Multiscale} deep
  feature embedding for ship detection in optical remote sensing imagery,''
  \emph{IEEE Transactions on Geoscience and Remote Sensing}, vol.~56, no.~12,
  pp. 7147--7161, 2018.

\bibitem{liu2018arbitrary}
W.~Liu, L.~Ma, and H.~Chen, ``Arbitrary-oriented ship detection framework in
  optical remote-sensing images,'' \emph{IEEE Geoscience and Remote Sensing
  Letters}, vol.~15, no.~6, pp. 937--941, 2018.

\bibitem{zhang2018toward}
Z.~Zhang, W.~Guo, S.~Zhu, and W.~Yu, ``Toward arbitrary-oriented ship detection
  with rotated region proposal and discrimination networks,'' \emph{IEEE
  Geoscience and Remote Sensing Letters}, vol.~15, no.~11, pp. 1745--1749,
  2018.

\bibitem{liu2017rotated}
Z.~Liu, J.~Hu, L.~Weng, and Y.~Yang, ``Rotated region based {CNN} for ship
  detection,'' in \emph{2017 IEEE International Conference on Image Processing
  (ICIP)}.\hskip 1em plus 0.5em minus 0.4em\relax IEEE, 2017, pp. 900--904.

\bibitem{ding2019learning}
J.~Ding, N.~Xue, Y.~Long, G.-S. Xia, and Q.~Lu, ``{Learning} {RoI} transformer
  for oriented object detection in aerial images,'' in \emph{Proceedings of the
  IEEE Conference on Computer Vision and Pattern Recognition}, 2019, pp.
  2849--2858.

\bibitem{li2020novel}
L.~Li, Z.~Zhou, B.~Wang, L.~Miao, and H.~Zong, ``A novel {CNN}-based method for
  accurate ship detection in {HR} optical remote sensing images via rotated
  bounding box,'' \emph{IEEE Transactions on Geoscience and Remote Sensing},
  2020.

\bibitem{liao2018rotation}
M.~Liao, Z.~Zhu, B.~Shi, G.-s. Xia, and X.~Bai, ``Rotation-sensitive regression
  for oriented scene text detection,'' in \emph{Proceedings of the IEEE
  conference on computer vision and pattern recognition}, 2018, pp. 5909--5918.

\bibitem{fu2020rotation}
K.~Fu, Z.~Chang, Y.~Zhang, G.~Xu, K.~Zhang, and X.~Sun, ``Rotation-aware and
  multi-scale convolutional neural network for object detection in remote
  sensing images,'' \emph{ISPRS Journal of Photogrammetry and Remote Sensing},
  vol. 161, pp. 294--308, 2020.

\bibitem{cheng2016learning}
G.~Cheng, P.~Zhou, and J.~Han, ``Learning rotation-invariant convolutional
  neural networks for object detection in {VHR} optical remote sensing
  images,'' \emph{IEEE Transactions on Geoscience and Remote Sensing}, vol.~54,
  no.~12, pp. 7405--7415, 2016.

\bibitem{zhou2017oriented}
Y.~Zhou, Q.~Ye, Q.~Qiu, and J.~Jiao, ``Oriented response networks,'' in
  \emph{Proceedings of the IEEE Conference on Computer Vision and Pattern
  Recognition}, 2017, pp. 519--528.

\bibitem{deng2018multi}
Z.~Deng, H.~Sun, S.~Zhou, J.~Zhao, L.~Lei, and H.~Zou, ``Multi-scale object
  detection in remote sensing imagery with convolutional neural networks,''
  \emph{ISPRS journal of photogrammetry and remote sensing}, vol. 145, pp.
  3--22, 2018.

\bibitem{wang2019fmssd}
P.~Wang, X.~Sun, W.~Diao, and K.~Fu, ``Fmssd: {Feature-merged} single-shot
  detection for multiscale objects in large-scale remote sensing imagery,''
  \emph{IEEE Transactions on Geoscience and Remote Sensing}, vol.~58, no.~5,
  pp. 3377--3390, 2019.

\bibitem{li2016novel}
S.~Li, Z.~Zhou, B.~Wang, and F.~Wu, ``A novel inshore ship detection via ship
  head classification and body boundary determination,'' \emph{IEEE geoscience
  and remote sensing letters}, vol.~13, no.~12, pp. 1920--1924, 2016.

\bibitem{girshick2014rich}
R.~Girshick, J.~Donahue, T.~Darrell, and J.~Malik, ``Rich feature hierarchies
  for accurate object detection and semantic segmentation,'' in
  \emph{Proceedings of the IEEE conference on computer vision and pattern
  recognition}, 2014, pp. 580--587.

\bibitem{girshick2015fast}
R.~Girshick, ``Fast {R-CNN},'' in \emph{Proceedings of the IEEE international
  conference on computer vision}, 2015, pp. 1440--1448.

\bibitem{ren2015faster}
S.~Ren, K.~He, R.~Girshick, and J.~Sun, ``Faster {R-CNN}: Towards real-time
  object detection with region proposal networks,'' in \emph{Advances in neural
  information processing systems}, 2015, pp. 91--99.

\bibitem{redmon2016you}
J.~Redmon, S.~Divvala, R.~Girshick, and A.~Farhadi, ``You only look once:
  {Unified}, real-time object detection,'' in \emph{Proceedings of the IEEE
  conference on computer vision and pattern recognition}, 2016, pp. 779--788.

\bibitem{redmon2017yolo9000}
J.~Redmon and A.~Farhadi, ``{YOLO9000}: {Better}, faster, stronger,'' in
  \emph{Proceedings of the IEEE conference on computer vision and pattern
  recognition}, 2017, pp. 7263--7271.

\bibitem{redmon2018yolov3}
J.~Redmon and A.~Farhadi, ``{YOLOv3}: {An} incremental improvement,''
  \emph{arXiv preprint arXiv:1804.02767}, 2018.

\bibitem{liu2016ssd}
W.~Liu, D.~Anguelov, D.~Erhan, C.~Szegedy, S.~Reed, C.-Y. Fu, and A.~C. Berg,
  ``{SSD}: {Single} shot multibox detector,'' in \emph{European conference on
  computer vision}.\hskip 1em plus 0.5em minus 0.4em\relax Springer, 2016, pp.
  21--37.

\bibitem{zhang2016weakly}
F.~Zhang, B.~Du, L.~Zhang, and M.~Xu, ``Weakly supervised learning based on
  coupled convolutional neural networks for aircraft detection,'' \emph{IEEE
  Transactions on Geoscience and Remote Sensing}, vol.~54, no.~9, pp.
  5553--5563, 2016.

\bibitem{yang2019r3det}
X.~Yang, Q.~Liu, J.~Yan, A.~Li, Z.~Zhang, and G.~Yu, ``R3det: {Refined}
  single-stage detector with feature refinement for rotating object,''
  \emph{arXiv preprint arXiv:1908.05612}, 2019.

\bibitem{zhang2019cad}
G.~Zhang, S.~Lu, and W.~Zhang, ``{CAD-Net}: {A} context-aware detection network
  for objects in remote sensing imagery,'' \emph{IEEE Transactions on
  Geoscience and Remote Sensing}, vol.~57, no.~12, pp. 10\,015--10\,024, 2019.

\bibitem{lin2017feature}
T.-Y. Lin, P.~Doll{\'a}r, R.~Girshick, K.~He, B.~Hariharan, and S.~Belongie,
  ``Feature pyramid networks for object detection,'' in \emph{Proceedings of
  the IEEE conference on computer vision and pattern recognition}, 2017, pp.
  2117--2125.

\bibitem{song2020revisiting}
G.~Song, Y.~Liu, and X.~Wang, ``Revisiting the sibling head in object
  detector,'' in \emph{Proceedings of the IEEE/CVF Conference on Computer
  Vision and Pattern Recognition}, 2020, pp. 11\,563--11\,572.

\bibitem{tian2019cascaded}
Z.~Tian, W.~Wang, R.~Zhan, Z.~He, J.~Zhang, and Z.~Zhuang, ``Cascaded detection
  framework based on a novel backbone network and feature fusion,'' \emph{IEEE
  Journal of Selected Topics in Applied Earth Observations and Remote Sensing},
  vol.~12, no.~9, pp. 3480--3491, 2019.

\bibitem{he2019bounding}
Y.~He, C.~Zhu, J.~Wang, M.~Savvides, and X.~Zhang, ``Bounding box regression
  with uncertainty for accurate object detection,'' in \emph{Proceedings of the
  IEEE Conference on Computer Vision and Pattern Recognition}, 2019, pp.
  2888--2897.

\bibitem{choi2019gaussian}
J.~Choi, D.~Chun, H.~Kim, and H.-J. Lee, ``Gaussian yolov3: {An} accurate and
  fast object detector using localization uncertainty for autonomous driving,''
  in \emph{Proceedings of the IEEE International Conference on Computer
  Vision}, 2019, pp. 502--511.

\bibitem{jiang2018acquisition}
B.~Jiang, R.~Luo, J.~Mao, T.~Xiao, and Y.~Jiang, ``Acquisition of localization
  confidence for accurate object detection,'' in \emph{Proceedings of the
  European Conference on Computer Vision (ECCV)}, 2018, pp. 784--799.

\bibitem{feng2018towards}
D.~Feng, L.~Rosenbaum, and K.~Dietmayer, ``Towards safe autonomous driving:
  {Capture} uncertainty in the deep neural network for lidar 3d vehicle
  detection,'' in \emph{2018 21st International Conference on Intelligent
  Transportation Systems (ITSC)}.\hskip 1em plus 0.5em minus 0.4em\relax IEEE,
  2018, pp. 3266--3273.

\bibitem{ming2020dynamic}
Q.~Ming, Z.~Zhou, L.~Miao, H.~Zhang, and L.~Li, ``Dynamic anchor learning for
  arbitrary-oriented object detection,'' \emph{arXiv preprint
  arXiv:2012.04150}, 2020.

\bibitem{lin2017focal}
T.-Y. Lin, P.~Goyal, R.~Girshick, K.~He, and P.~Doll{\'a}r, ``Focal loss for
  dense object detection,'' in \emph{Proceedings of the IEEE international
  conference on computer vision}, 2017, pp. 2980--2988.

\bibitem{liu2017high}
Z.~Liu, L.~Yuan, L.~Weng, and Y.~Yang, ``A high resolution optical satellite
  image dataset for ship recognition and some new baselines,'' in
  \emph{Proceedings of the International Conference on Pattern Recognition
  Applications and Methods}, vol.~2, 2017, pp. 324--331.

\bibitem{xia2018dota}
G.-S. Xia, X.~Bai, J.~Ding, Z.~Zhu, S.~Belongie, J.~Luo, M.~Datcu, M.~Pelillo,
  and L.~Zhang, ``{DOTA}: {A} large-scale dataset for object detection in
  aerial images,'' in \emph{Proceedings of the IEEE Conference on Computer
  Vision and Pattern Recognition}, 2018, pp. 3974--3983.

\bibitem{zhu2015orientation}
H.~Zhu, X.~Chen, W.~Dai, K.~Fu, Q.~Ye, and J.~Jiao, ``Orientation robust object
  detection in aerial images using deep convolutional neural network,'' in
  \emph{2015 IEEE International Conference on Image Processing (ICIP)}.\hskip
  1em plus 0.5em minus 0.4em\relax IEEE, 2015, pp. 3735--3739.

\bibitem{he2016deep}
K.~He, X.~Zhang, S.~Ren, and J.~Sun, ``Deep residual learning for image
  recognition,'' in \emph{Proceedings of the IEEE conference on computer vision
  and pattern recognition}, 2016, pp. 770--778.

\bibitem{everingham2010pascal}
M.~Everingham, L.~Van~Gool, C.~K. Williams, J.~Winn, and A.~Zisserman, ``The
  pascal visual object classes (voc) challenge,'' \emph{International journal
  of computer vision}, vol.~88, no.~2, pp. 303--338, 2010.

\bibitem{jiang2017r2cnn}
Y.~Jiang, X.~Zhu, X.~Wang, S.~Yang, W.~Li, H.~Wang, P.~Fu, and Z.~Luo,
  ``{R2CNN}: {Rotational} region cnn for orientation robust scene text
  detection,'' \emph{arXiv preprint arXiv:1706.09579}, 2017.

\bibitem{ma2018arbitrary}
J.~Ma, W.~Shao, H.~Ye, L.~Wang, H.~Wang, Y.~Zheng, and X.~Xue,
  ``Arbitrary-oriented scene text detection via rotation proposals,''
  \emph{IEEE Transactions on Multimedia}, vol.~20, no.~11, pp. 3111--3122,
  2018.

\bibitem{xu2020gliding}
Y.~Xu, M.~Fu, Q.~Wang, Y.~Wang, K.~Chen, G.-S. Xia, and X.~Bai, ``Gliding
  vertex on the horizontal bounding box for multi-oriented object detection,''
  \emph{IEEE Transactions on Pattern Analysis and Machine Intelligence}, 2020.

\bibitem{yang2018automatic}
X.~Yang, H.~Sun, K.~Fu, J.~Yang, X.~Sun, M.~Yan, and Z.~Guo, ``Automatic ship
  detection in remote sensing images from google earth of complex scenes based
  on multiscale rotation dense feature pyramid networks,'' \emph{Remote
  Sensing}, vol.~10, no.~1, p. 132, 2018.

\bibitem{azimi2018towards}
S.~M. Azimi, E.~Vig, R.~Bahmanyar, M.~K{\"o}rner, and P.~Reinartz, ``Towards
  multi-class object detection in unconstrained remote sensing imagery,'' in
  \emph{Asian Conference on Computer Vision}.\hskip 1em plus 0.5em minus
  0.4em\relax Springer, 2018, pp. 150--165.

\bibitem{pan2020dynamic}
X.~Pan, Y.~Ren, K.~Sheng, W.~Dong, H.~Yuan, X.~Guo, C.~Ma, and C.~Xu, ``Dynamic
  refinement network for oriented and densely packed object detection,'' in
  \emph{Proceedings of the IEEE/CVF Conference on Computer Vision and Pattern
  Recognition}, 2020, pp. 11\,207--11\,216.

\bibitem{wei2020oriented}
H.~Wei, Y.~Zhang, Z.~Chang, H.~Li, H.~Wang, and X.~Sun, ``Oriented objects as
  pairs of middle lines,'' \emph{ISPRS Journal of Photogrammetry and Remote
  Sensing}, vol. 169, pp. 268--279, 2020.

\bibitem{yang2019scrdet}
X.~Yang, J.~Yang, J.~Yan, Y.~Zhang, T.~Zhang, Z.~Guo, X.~Sun, and K.~Fu,
  ``Scrdet: {Towards} more robust detection for small, cluttered and rotated
  objects,'' in \emph{Proceedings of the IEEE International Conference on
  Computer Vision}, 2019, pp. 8232--8241.

\end{thebibliography}
\end{document}